\RequirePackage{amsmath}
\documentclass[runningheads]{llncs}
\usepackage{amsfonts}
\usepackage{hyperref}
\usepackage{graphicx}
\usepackage{comment}
\usepackage{amssymb}
\usepackage{bm}
\usepackage{breakcites}

\usepackage{booktabs}
\usepackage{array}
\usepackage{graphicx}
\usepackage{multirow}

\usepackage{floatrow}
\DeclareFloatFont{tiny}{\scriptsize}
\makeatletter
\def\th@plain{%
  \thm@notefont{}
  \itshape 
}
\def\th@definition{%
  \thm@notefont{}
  \normalfont 
}
\makeatother
\newcommand{\tr}[1]{\langle #1 \rangle}

\usepackage{cleveref}

\usepackage{algpseudocode,algorithm,algorithmicx}

\usepackage{booktabs}

\usepackage{tikz}
\usetikzlibrary{automata,shapes.multipart} 
\usetikzlibrary{arrows,petri}
\usetikzlibrary{arrows}
\usetikzlibrary{shapes,backgrounds}
\usetikzlibrary{arrows,shapes,backgrounds,patterns,decorations.pathreplacing,decorations.pathmorphing,decorations.markings,shadows,shapes.misc,calc,positioning}
\usetikzlibrary {graphs}

\usepackage[colorinlistoftodos]{todonotes}
\definecolor{forestgreen}{rgb}{0.13,0.55,0.13}
\definecolor{amber}{rgb}{1.0,0.49,0}
\definecolor{cargreen}{rgb}{0.0,0.8,0.6}
\definecolor{aureolin}{rgb}{0.54,0.17,0.89}
\definecolor{bured}{rgb}{0.8,0.0,0.0}

\colorlet{orange}{red!70!yellow}
\colorlet{vert}{green!70!blue}
\colorlet{mauve}{blue!70!red}
\colorlet{brouge}{red!70!blue}
\colorlet{frouge}{red!40!black}

\usepackage[T1]{fontenc}
\usepackage{graphicx}
\usepackage{amsthm}
\theoremstyle{definition}
\newtheorem{procedure}{Procedure}
\newtheorem{myremark}{Remark}

\allowdisplaybreaks

\begin{document}
%
\title{Probabilistic Process Discovery with Stochastic Process Trees}

%
%
\author{
Andr\'as Horv\'ath\inst{1}
\and
Paolo Ballarini\inst{2}
\and
Pierre Cry\inst{2}
}
%
%
\institute{
Università di Torino, Torino, Italy\\
\email{horvath@di.unito.it}
\and
Université Paris Saclay, Lab. MICS, CentraleSupélec,  Gif-sur-Yvette, France 
\email{\{paolo.ballarini,pierre.cry\}@centralesupelec.fr}
}
\maketitle              
\begin{abstract}
In order to obtain a stochastic model that accounts for the stochastic aspects of  the dynamics of a business process, usually the following steps are taken. Given an event log, a process tree is obtained through a process discovery algorithm, i.e., a process tree that is aimed at reproducing, as accurately as possible,  the language of the log. The process tree is then transformed into a Petri net that generates the same set of sequences as the process tree. In order to capture the frequency of the sequences in the event log, weights are assigned to the transitions of the Petri net, resulting in a stochastic Petri net with a stochastic language in which each sequence is associated with a probability.

In this paper we show that this procedure has unfavorable properties. First, the weights assigned to the
transitions of the Petri net have an unclear role in the
resulting stochastic language. We will show that a weight can
have multiple, ambiguous impact on the probability of the
sequences generated by the Petri net. Second, a number of
different Petri nets with different number of transitions can
correspond to the same process tree. This means that the number
of parameters (the number of weights) that determines the
stochastic language is not well-defined.

In order to avoid these ambiguities, in this paper, we propose to
add stochasticity directly to process trees. The result is a new
formalism, called stochastic process trees, in which the number of
parameters and their role in the associated stochastic language
is clear and well-defined.
\end{abstract}
\section{Introduction}
\label{sec:intro}

\emph{Process mining}~\cite{10.5555/2948762}  is concerned with the    \emph{discovery} of  formal models that adequately reproduce the dynamics of a  business process. Model discovery exploit observations of the considered process, normally consisting  of sequences of events, called \emph{traces}, accumulated in the form of \emph{event logs}. A \emph{trace}, consists in a sequence of (possibly timestamped) activities that represent one observed execution of the process. 
Classic process mining algorithms, e.g., ~\cite{1316839,DBLP:conf/apn/LeemansFA13,10.1007/s10115-018-1214-x}, are aimed at capturing the workflow aspects of the considered process, that is,  at extracting a so-called \emph{workflow} model, normally in the form of a structured Petri net, capable of mimicking  the \emph{log's language} (i.e., the model should be able  to replay the traces of the log).  
The derivation of a workflow model from a log entails  ensuring that 
the correct  sequencing of activities (e.g., whether an activity $b$ directly follows an activity $a$), the correct concurrency between activities (e.g., whether an activity $a$ may occur concurrently to an activity $b$)  as well as  the  correct competition  (e.g., whether the occurrence of an activity $a$ may exclude that  of an activity $b$) as exhibited by the traces in the log. 
The quality of discovery algorithms is assessed by means of a number of  indicators  including \emph{fitness}, which measures how much of the language of the log (i.e., the set of traces in the log)  is reproduced by the discovered model, and \emph{precision}, which, conversely,  measures how much of the language of the model (the traces issued by the model) is contained in the log. 

Some discovery algorithms  output models using  formalisms other than the Petri net one. This is the case, for example, with algorithms of the inductive miner (IM) family~\cite{DBLP:conf/apn/LeemansFA13}, a popular kind of discovery methods 
which  enjoy several nice properties including \emph{perfect fitness}, that is,  guaranteeing that the discovered model reproduces all traces of the log. 
IM algorithms discover a \emph{process tree} model  from an event log, that is,  
a tree  obtained by combining a number of ``workflow'' operators (i.e., sequence, choice, parallel, loop) applied to the process activities appearing as leaves. The resulting tree is suitable for a translation into a hierarchically structured Petri net whose blocks corresponds to the subtrees of the obtained tree. 

\medskip
\noindent
{\it The stochastic process mining problem.}
Traces of an event log are often equipped with a multiplicity (also called \emph{frequency}) which indicates how many times  each unique sequence of actions has been detected while observing  the process. Through a simple normalization step, trace multiplicity   allows to account for  the likelihood that the process exhibits the  specific behavior corresponding to a  trace, hence each trace of the log is characterized by a probability value and therefore the log  form a so-called \emph{stochastic language}.
This leads to the stochastic extension of the process mining problem, namely: by taking into account trace multiplicity we aim to extract a stochastic model  whose  \emph{stochastic language}  resembles that of the log. 
Within the  still relatively emergent literature~\cite{5f0e8dd04572478fb450eefe4211d69f,DBLP:journals/corr/abs-2406-10817,DBLP:conf/icpm/BurkeLW20,DBLP:conf/apn/BurkeLW21,10.1007/11494744_25}, mining of a stochastic model is mostly achieved \emph{indirectly}, that is, first a non-stochastic, workflow model is extracted from the log using standard discovery algorithms (hence disregarding trace multiplicity),  then the obtained model is converted into a stochastic one by associating \emph{weight} parameters to each event. 
This leads to a \emph{parameter optimization} problem aimed at identifying adequate values for the model parameters so that the resulting  stochastic language of the model  is similar to that of the log, a problem whose complexity  is  dependent on the number of parameters.

\medskip
\noindent
{\it Problem faced.}
To the best of our knowledge existing stochastic discovery approaches rely all on   Petri nets as modeling formalism. We argue that, although effective, the use of  Petri net  models, may overly complicate the search for optimal parameters, specially whenever the employed discovery algorithm is likely to generate a Petri net with many (redundant) transitions, as it is the case with the IM algorithm. 

\medskip
\noindent
{\it Our contribution.}
As a workaround  we propose to exploit the more succinct process tree formalism directly as a means to reduce the dimensionality of  parameter search within the stochastic process mining problem. To this aim we introduce the  notion of \emph{stochastic process tree} (SPT), essentially by extending the standard process tree (PT) operators~\cite{DBLP:conf/apn/LeemansFA13} with probabilistic arguments that associate each trace that can be generated by the tree with a probability, resulting in a stochastic language.


This allows us to adapt  stochastic process discovery so that it operates directly on SPTs, simply obtained by associating probabilistic parameters to the nodes of a PT issued  by a standard discovery algorithm such as IM, rather than operating on the dimensionally  much larger  Petri nets obtained by conversion of the discovered PT. 

\medskip
\noindent
{\it Paper structure.} 
The paper is organized as follows: in Section~\ref{sec:prelim} we introduce some preliminaries we rely upon throughout the manuscript, in Section~\ref{sec:example} we discuss an example which motivates the proposed approach by highlighting the presence of redundant (uninfluential)  parameters in Petri nets discovered through standard process discovery methods. In Section~\ref{sec:method} we introduce the novel stochastic process tree formalism, including its syntax, its semantics as well as a procedure for sampling traces from the stochastic language it represents. In Section~\ref{sec:exper} we present the results of experiments obtained through a Python encoded prototypical implementation of stochastic process tree based approach for process discovery. We wrap up the manuscript in Section~\ref{sec:conclusion} with a few summary remarks which include also perspectives about future developments.


\section{Preliminaries}
\label{sec:prelim}
We briefly summarise a number of notions and related notations that constitute the basis for formalising the process mining problem.

\noindent
\textit{Alphabets, traces,  languages and logs}.
We let  $\Sigma$ denote the alphabet of the process activities (we use  letters to denote activities, e.g., $\Sigma=\{a,b,c\}$) and  $\Sigma^*$  the set of possibly infinite traces (words, sequences) composed of activities in $\Sigma$ where $\epsilon\in\Sigma^*$ represents the empty trace. We let $S\subset \Sigma^*$  denote a generic set of traces built on  alphabet $\Sigma$ and $\sigma$ a trace in $S$, for example, $\sigma_1=\langle a,b,b,a,c\rangle\in\{a,b,c\}^*$. The length of a trace $\sigma$ is denoted by $|\sigma|$ while its $k$-th activity by $\sigma[k]$.

For any two traces $\sigma_1,\sigma_2\in S$ we denote $\sigma_1. \sigma_2$ the concatenation of  $\sigma_2$ to $\sigma_1$, e.g., with $\sigma_2=\langle e,f\rangle$ we have $\sigma_1.\sigma_2=\langle a,b,b,a,c,e,f\rangle$. Trace concatenation extends naturally  to set of traces. For two sets of traces $S_1,S_2$ we denote by $S_1\bigodot S_2$ the set consisting of traces resulting by concatenating any trace of $S_1$ with any trace of $S_2$, e.g., with $S_1=\{\langle a,b,c,\rangle\, \langle c,d\rangle\}$ and $S_2=\{\langle b,d\rangle\, \langle e,e\rangle\}$ we have $S_1\bigodot S_2=\{ \langle a,b,c,b,d\rangle\, \langle a,b,c,e,e\rangle\, \langle c,d,b,d\rangle, \langle c,d,e,e\rangle\}$.

All possible interleavings (parallel executions) of two traces $\sigma_1,\sigma_2$ will be denoted by $\sigma_1\diamondsuit \sigma_2$. For example, with $\sigma_1=\tr{a,b},\sigma_2=\tr{c}$, we have $\sigma_1\diamondsuit \sigma_2=\{\tr{a,b,c},$ $\tr{a,c,b},$ $\tr{c,a,b}\}$. As concatenation, also interleaving, $\diamondsuit$, is extended to set of traces in the natural way. 
 
 An event log $E$ is a multi-set of traces built on an alphabet $\Sigma$, i.e., $E\in Bag(\Sigma^*)$. Given a trace $\sigma \in E$ we denote by $f(E,\sigma)$ its multiplicity (i.e., its frequency in $E$).


\noindent
\textit{Stochastic language}. A stochastic language (SL) over the traces built on an alphabet $\Sigma$ is a function $L:\Sigma^*\to[0,1]$ 
that provides the probability of the traces 
such that $\sum_{\sigma\in\Sigma^*} L(\sigma) = 1$. 
The stochastic language induced by an event log $E$ is straightforwardly obtained by computing the probability of the traces as $p(E,\sigma)=f(E,\sigma)/\sum_{\sigma\in Supp(E)} f(E,\sigma)$ where $Supp(E)$ denotes the set of unique traces in $E$, i.e., its support. 
Given $n$ SLs $L_i$ 
and $n$ weights $p_i\in[0,1]$ with $\sum_{i=1}^1 p_i =1$, the mixture distribution~\cite{pearson1894} of $L_i$ w.r.t. $p_i$ is the distribution in which $L(\sigma)=\sum_{i=1}^n p_i\cdot L_i(\sigma)$. 
\noindent
\textit{Earth Movers Distance}.
To measure the resemblance between two stochastic languages we refer to the Earth Movers Distance (EMD)~\cite{rubner1998signature}, whose basic principle is to quantify the minimal amount of \emph{work} needed to redistribute the probability mass distribution (pmd) of one of the two languages in such a way that it becomes identical to the pmd of the other language\footnote{In our case EMD is calculated based on the so-called the Levenshtein distance which measures the distance between two
words as the minimum \emph{alignment}~\cite{https://doi.org/10.1002/widm.1045}, that is, the minimum number of
single-character edits (insertion, deletion or substitution of an action)
needed to change one word (trace) into the other.}. 
EMD distance equal to 0 means that the two distributions are equal and the distance is normalized in such a way that the maximal distance is 1.
Whereas for small languages the assessment of EMD   is computationally feasible, it becomes computationally hard when it comes with infinite (or very large) languages. 
Therefore   approximations of the EMD based on finite subsets of the compared languages, such as the \emph{truncated} EMD (tMED)~\cite{DBLP:conf/icpm/BurkeLW20}, are commonly considered for stochastic conformance checking~\cite{10.1007/978-3-030-26643-1_8}. 
In this work we consider the so-called \emph{restricted} EMD (rEMD)~\cite{DBLP:journals/corr/abs-2406-10817} which is suitable to assess the resemblance between a finite language $L_1$ and an infinite one $L_2$ that contains it (i.e., $L_1\subset L_2$). In essence, with rEMD the distance is computed only with respect to the words the two languages share, i.e., the words in $L_1$.

\noindent
\textit{Petri nets}. A labeled Petri net (PN) model is a tuple $N=(P,T,F,\Sigma,\lambda,M_0)$, where $P$ is the set of places, $T$ the set of transitions, 
$F\subseteq (P\times T)\cup(T\times P)$ is the set of directed arcs, 
$\lambda: T\to (\Sigma\cup\{\tau\})$ associates each transition with an action ($\tau$ being the silent action) and $M_0:P\to\mathbb{N}$ is the initial marking. 
Given a marking $M$, $M(p)$ is the number of tokens contained in place $p$ in marking $M$. For simplicity we denote markings as multi-sets over $P$, for example for a net with places $P=\{p_1,p_2,p_3\}$, $M_1=2p_1+3p_2$ is the marking where $p_1$  and $p_2$  contains, respectively, 2 and 3 tokens (and $p_3$ none), while $M_2=p_3$ is the marking where $p_3$ contains 1 token.

For a PN $N$, we denote by $RG(N)=(S,A)$ the reachability graph where $S=RS(N)$ is the reachability set (the set of  markings reachable from the initial one) and $A\subseteq RS(N)\times RS(N)\times T$  is the set of arcs whose elements $(M,M',t)\in A$ are such that $M[t\rangle M'$, that is, $M'$ is reached from $M$ by firing of $t$. 
For any  node $x\in P\cup T$ we denote $\bullet x=\{x'|F(x',x)>0\}$, respectively $x\bullet =\{x'|F(x,x')>0\}$,  the set of input, respectively output, nodes of $x$.


\noindent
\textit{Workflow nets}. A workflow net (WN) is a 1-safe\footnote{In any marking each place may contain at most 1 token.} Petri net $N=(P,T,F,\Sigma,\lambda,M_0)$ with the following structural constraints: 1) there exists a unique place, denoted \emph{source} with no incoming transition ($\bullet source =\emptyset$) and a unique place denoted \emph{sink} with no outgoing transitions ($sink\bullet =\emptyset$); 2) the initial marking is $M_0=source)$ and 3) the net 
$\bar{N}=(P,T\cup\{\bar{t}\},F\cup\{(sink,\bar{t}),(\bar{t},source)\},\Sigma\cup\{\tau\},\lambda\cup\{(\bar{t},\tau)\},M_0)$ obtained by adding a transition $\bar{t}$ connecting the $sink$ to the $source$ place  is strongly connected~\footnote{The latter constraint is also often formulated  as \emph{``each node of the net must be on a path connecting the source and sink places''}.}.

\noindent
\textit{Block-structured workflow nets}. A block-structured workflow net is a hierarchically structured workflow net, that is, a workflow net that is recursively decomposed into parts each of which is a workflow net (i.e., equipped with a single source and a single sink place).  

\begin{example}
Figure~\ref{fig:wnmapping} shows an example of block-structured WN. 
Silent transitions (labeled with $\tau$) are black filled while all other transitions correspond to a given activity of the log (the label is depicted inside the transition). 
\end{example}

\noindent
\textit{Process trees}. Process trees~\cite{10.1007/978-3-642-34044-4_3} (PT) is a formalism used to obtain a hierarchical representation of a set of traces (typically the traces contained in an event log). The leaves of a PT correspond to activities while the internal nodes correspond to operators through which the languages of the corresponding sub-trees are combined. 
Four operators are commonly considered to build process trees, namely: \textit{sequence} ($\rightarrow$), \textit{choice} ($\times$), 
\textit{parallel} ($\land$) and  \textit{loop} ($\circlearrowleft$). 
In order to  introduce the notion of stochastic process tree (Section~\ref{sec:method}), we recall here the formal definition of process tree syntax and semantics (Definition~\ref{def:ptsyntax} and~\ref{def:ptsemanitcs}). 

\begin{definition}[Process tree] 
\label{def:ptsyntax}
Let $\Sigma$ be an alphabet of process activities and $\{\rightarrow, \times, \land, \circlearrowleft  \}$ the set of process tree operators. The set of process trees over $\Sigma$ is recursively defined as:
\begin{itemize}
    \item if $a \in\Sigma \cup \{\tau\}$ then $Q=a$ is a process tree,
    \item if $Q_1,Q_2,...,Q_n$ with $n \geq 2$ are process trees and $\oplus \in \{\rightarrow, \times, \land \}$ then $\oplus(Q_1,Q_2,...,Q_n)$ is a process tree,
    \item if $Q_1$ and $Q_2$ are process trees then $\circlearrowleft(Q_1,Q_2)$ is a process tree.
\end{itemize}
\end{definition}

\begin{definition}[Semantics of process trees]
\label{def:ptsemanitcs}
Let $Q$ be a process tree over the alphabet $\Sigma$. The language of $Q$, denoted by $\mathcal{L}(Q)$, is defined recursively as:
\begin{itemize}
\item if $Q=a$ with $a\in\Sigma$ then $\mathcal{L}(Q)=\{\tr{a}\}$,
\item if $Q=\tau$  then $\mathcal{L}(Q)=\{\tr{}\}$,
\item (sequence:) if $Q=\rightarrow(Q_1,Q_2,...,Q_n)$ then $\mathcal{L}(Q)=\bigodot_{i=1}^n \mathcal{L}(Q_i)$,
\item (choice:) if $Q=\times(Q_1,Q_2,...,Q_n)$ then $\mathcal{L}(Q)=\bigcup_{i=1}^n \mathcal{L}(Q_i)$,
\item (parallel:) if $Q=\land(Q_1,Q_2,...,Q_n)$ then $\mathcal{L}(Q)=
\diamondsuit_{i=1}^n \mathcal{L}(Q_i)$,
\item (loop:) if $Q=\circlearrowleft(Q_1,Q_2)$ then $\mathcal{L}(Q)=\{\sigma_1.\sigma_1'.\sigma_2.\sigma_2'.~...~.\sigma_{m-1}'.\sigma_m\big|m\geq 1,~ \forall i, 1\leq i \leq m,\sigma_i\in \mathcal{L}(Q_1),~ \forall j, 1\leq j \leq m-1, \sigma_j'\in \mathcal{L}(Q_2)\}$.
\end{itemize}
\end{definition}

\begin{myremark}
Note  that, differently from the original definition of PT semantics~\cite{10.1007/978-3-642-34044-4_3}, in Definition~\ref{def:ptsemanitcs}) we opted for a binary ($\circlearrowleft(Q_1,Q_2)$), rather than an $n$-ary ($\circlearrowleft(Q_1,Q_2,\dots ,Q_n)$) version  for the loop operator. We point out that this does not affect the expressiveness of the PT formalism as the following equivalence holds: $\circlearrowleft(Q_1,Q_2,\dots ,Q_n)=\circlearrowleft(Q_1,\times(Q_2,\ldots ,Q_n))$.
\end{myremark}

\begin{example}
Figure~\ref{fig:pt1} shows a PT representation of the event log 
$L_1=\{\langle a,c,d,e\rangle, \allowbreak \langle b,c,d,e\rangle,\langle c,b,d,f\rangle,\langle a,c,c,d,e\rangle, \langle c,b,d,f\rangle,\langle c,c,a,d,f\rangle\}$.
\begin{figure}
    \centering
    \resizebox{0.325\textwidth}{!}{
\begin{tikzpicture}[>=stealth, node distance=1.45cm, thick]

\tikzstyle{circle_node} = [circle, draw, minimum size=2em]
\tikzstyle{square_node} = [rectangle, draw, fill=red!20, minimum size=2em]

\node[circle_node] (Seq) {$\rightarrow$};
\node[circle_node] (Paral) [below left of=Seq] {$\land$};

\node[circle_node] (Choice1) [below left of=Paral] {$\times$};

\node[circle_node] (Choice2) [below right of=Seq] {$\times$};

\node[circle_node] (Loop) [right  of=Choice1] {$\circlearrowleft$};

\node[square_node] (A) [below left of=Choice1] {$a$};
\node[square_node] (B) [right  of=A] {$b$};
\node[square_node] (C) [below  of=Loop] {$c$};
\node[square_node] (Tau) [right  of=C] {$\tau$};

\node[square_node] (E) [below  of=Choice2] {$e$};
\node[square_node] (F) [right  of=E] {$f$};

\node[square_node] (D) [below  of=Seq] {$d$};

\draw[->] (Seq) -- (Paral);
\draw[->] (Seq) -- (Choice2);
\draw[->] (Paral) -- (Choice1);
\draw[->] (Paral) -- (Loop);
\draw[->] (Choice1) -- (A);
\draw[->] (Choice1) -- (B);
\draw[->] (Choice2) -- (E);
\draw[->] (Choice2) -- (F);

\draw[->] (Loop) -- (C);
\draw[->] (Loop) -- (Tau);

\draw[->] (Seq) -- (D);

\end{tikzpicture}
}
    \caption{A process tree corresponding to log $L_1$}
    \label{fig:pt1}
\end{figure}
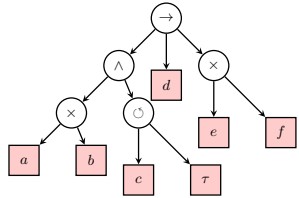
Looking at the traces in $L_1$, we observe the following characteristics: each trace contains action $d$, and  $d$ is always preceded by any combination of either $a$ or $b$ shuffled with either one or two consecutive occurrences of $c$. Also, $d$ is always followed by either $e$ or $f$. The corresponding PT is given by $Q_1=\rightarrow(\land(\times(a,b),\circlearrowleft(c,\tau)),d,\times(e,f))$. By application of the  semantics, it is straightforward to show that the log $L_1$ is contained in the language of the tree, i.e., $L_1\subset\mathcal{L}(Q_1)$. It is also evident that, since in the PT activity $c$ can be executed any number of times (due to the loop sub-tree $\circlearrowleft(c,\tau)$) , $\mathcal{L}(Q_1)$ contains traces that are not present in the event log.   
\end{example}

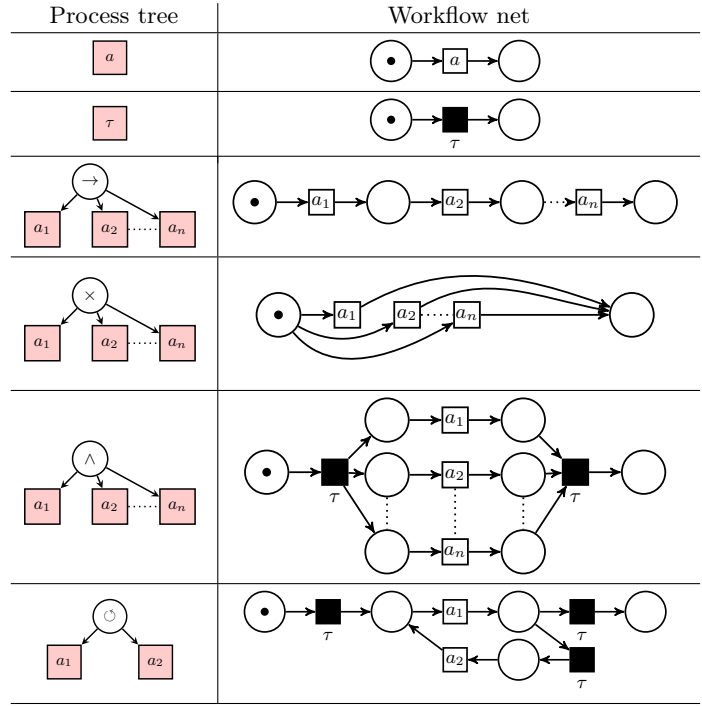
\begin{figure}[t]
    \centering
    \begin{tabular}{c|c}
    Process tree & Workflow net \\
    \hline
      \begin{tabular}{c}
      \resizebox{0.05\textwidth}{!}{
\begin{tikzpicture}[>=stealth, node distance=1.25cm, thick]

\tikzstyle{circle_node} = [circle, draw, minimum size=2em]
\tikzstyle{square_node} = [rectangle, draw, fill=red!20, minimum size=2em]

\node[square_node] (A) [] {$a$};

\end{tikzpicture}
}
      \end{tabular}& 
      \begin{tabular}{c}\\[-3mm]
      \resizebox{0.2\textwidth}{!}{
\begin{tikzpicture}[node distance=.5cm,>=stealth',thick]

\node[place,tokens=1] (p1) {};

\node[transition] (t1) [right=of p1, label=center:$a$] {};
\node[place] (p2) [right=of t1] {};

\draw[->] (p1) -- (t1);
\draw[->] (t1) -- (p2);

\end{tikzpicture}
} 
      \end{tabular}\\
      \hline
      \begin{tabular}{c}
      \resizebox{0.05\textwidth}{!}{
\begin{tikzpicture}[>=stealth, node distance=1.25cm, thick]

\tikzstyle{circle_node} = [circle, draw, minimum size=2em]
\tikzstyle{square_node} = [rectangle, draw, fill=red!20, minimum size=2em]

\node[square_node] (A) [] {$\tau$};

\end{tikzpicture}
}
      \end{tabular}& 
      \begin{tabular}{c}\\[-3mm]\resizebox{0.2\textwidth}{!}{
\begin{tikzpicture}[node distance=.5cm,>=stealth',thick]

\node[place,tokens=1] (p1) {};

\node[transition,fill=black] (t1) [right=of p1, label=below:$\tau$] {};
\node[place] (p2) [right=of t1] {};

\draw[->] (p1) -- (t1);
\draw[->] (t1) -- (p2);

\end{tikzpicture}
} 
      \end{tabular}\\[-1mm]
    \hline
      \begin{tabular}{c}\\[-3mm]
      \resizebox{0.2\textwidth}{!}{

\begin{tikzpicture}[>=stealth, node distance=1.25cm, thick]

\tikzstyle{circle_node} = [circle, draw, minimum size=2em]
\tikzstyle{square_node} = [rectangle, draw, fill=red!20, minimum size=2em]

\node[circle_node] (Root) {$\rightarrow$};
\node[square_node] (A1) [below left of=Root] {$a_1$};
\node[square_node] (A2) [right  of=A1] {$a_2$};
\node[square_node] (An) [right  of=A2] {$a_n$};

\draw[->] (Root) -- (A1);
\draw[->] (Root) -- (A2);
\draw[->] (Root) -- (An);
\draw[dotted] (A2) -- (An);

\end{tikzpicture}

}
      \end{tabular}& 
      \begin{tabular}{c}\resizebox{0.5\textwidth}{!}{
\begin{tikzpicture}[node distance=.5cm,>=stealth',thick]

\node[place,tokens=1] (p1) {};

\node[transition] (t1) [right=of p1, label=center:$a_1$] {};
\node[place] (p2) [right=of t1] {};

\node[transition] (t2) [right=of p2, label=center:$a_2$] {};

\node[place] (p3) [right=of t2] {};

\node[transition] (t3) [right=of p3, label=center:$a_n$] {};

\node[place] (p4) [right=of t3] {};

\draw[->] (p1) -- (t1);
\draw[->] (t1) -- (p2);
\draw[->] (p2) -- (t2);
\draw[->] (t2) -- (p3);
\draw[->,dotted] (p3) -- (t3);
\draw[->] (t3) -- (p4);

\end{tikzpicture}
} 
      \end{tabular}\\
    \hline
      \begin{tabular}{c}
      \resizebox{0.2\textwidth}{!}{
\begin{tikzpicture}[>=stealth, node distance=1.25cm, thick]

\tikzstyle{circle_node} = [circle, draw, minimum size=2em]
\tikzstyle{square_node} = [rectangle, draw, fill=red!20, minimum size=2em]

\node[circle_node] (Root) {$\times$};
\node[square_node] (A1) [below left of=Root] {$a_1$};
\node[square_node] (A2) [right  of=A1] {$a_2$};
\node[square_node] (An) [right  of=A2] {$a_n$};

\draw[->] (Root) -- (A1);
\draw[->] (Root) -- (A2);
\draw[->] (Root) -- (An);
\draw[dotted] (A2) -- (An);

\end{tikzpicture}
}
      \end{tabular}& 
      \begin{tabular}{c}\resizebox{0.45\textwidth}{!}{
\begin{tikzpicture}[node distance=.5cm,>=stealth',thick]

\node[place, tokens=1] (p1) {};
\node[place] (p4) [right=of p3] {};

\node[transition] (t1) [right=of p1, label=center:$a_1$] {};
\node[transition] (t2) [right=of t1, label=center:$a_2$] {};
\node[transition] (t3) [right=of t2, label=center:$a_n$] {};

\draw[->] (p1) -- (t1);
\draw[->] (p1) to [out=-30,in=210]  (t2);
\draw[->] (p1) to [out=-50,in=210]  (t3);
\draw[->] (t1) to [out=30,in=160] (p4);
\draw[->] (t2) to [out=30,in=170] (p4);
\draw[->] (t3) -- (p4);
\draw[-, dotted] (t2) -- (t3);


\end{tikzpicture}
} 
      \end{tabular}\\
      
      \hline
    \begin{tabular}{c}
      \resizebox{0.2\textwidth}{!}{
\begin{tikzpicture}[>=stealth, node distance=1.25cm, thick]

\tikzstyle{circle_node} = [circle, draw, minimum size=2em]
\tikzstyle{square_node} = [rectangle, draw, fill=red!20, minimum size=2em]

\node[circle_node] (Root) {$\land$};
\node[square_node] (A1) [below left of=Root] {$a_1$};
\node[square_node] (A2) [right  of=A1] {$a_2$};
\node[square_node] (An) [right  of=A2] {$a_n$};

\draw[->] (Root) -- (A1);
\draw[->] (Root) -- (A2);
\draw[->] (Root) -- (An);
\draw[dotted] (A2) -- (An);

\end{tikzpicture}

}
      \end{tabular}& 
      \begin{tabular}{c}\\[-3mm]\resizebox{0.475\textwidth}{!}{
\begin{tikzpicture}[node distance=.5cm,>=stealth',thick]


\node[place,tokens=1] (p0) {};
\node[transition] (t0) [right=of p0, fill=black,label=below:$\tau$] {};

\node[place] (p1) [above right =of  t0] {};
\node[place] (p2) [below = 0.15cm of p1] {};
\node[place] (pn) [below = 0.5cm of p2] {};

\node[transition] (t1) [right=of p1, label=center:$a_1$] {};
\node[place] (p1end) [right=of t1] {};

\node[transition] (t2) [right=of p2, label=center:$a_2$] {};
\node[place] (p2end) [right=of t2] {};

\node[transition] (tn) [right=of pn, label=center:$a_n$] {};
\node[place] (pnend) [right=of tn] {};



\node[transition] (tend) [below right=of p1end, fill=black,label=below:$\tau$] {};

\node[place] (pend) [right =of tend]{};

\draw[->] (p0) -- (t0);
\draw[->] (t0) -- (p1);
\draw[->] (t0) -- (p2);
\draw[->] (t0) -- (pn);
\draw[->] (p1) -- (t1);
\draw[->] (t1) -- (p1end);
\draw[->] (p2) -- (t2);
\draw[->] (pn) -- (tn);
\draw[->] (t2) -- (p2end);
\draw[->] (tn) -- (pnend);
\draw[->] (p1end) -- (tend);
\draw[->] (p2end) -- (tend);
\draw[->] (pnend) -- (tend);
\draw[->] (tend) -- (pend);
\draw[-,dotted] (p2) -- (pn);
\draw[-,dotted] (t2) -- (tn);
\draw[-,dotted] (p2end) -- (pnend);

\end{tikzpicture}
} 
      \end{tabular}\\
      \hline
    \begin{tabular}{c}
      \resizebox{0.15\textwidth}{!}{
\begin{tikzpicture}[>=stealth, node distance=1.25cm, thick]

\tikzstyle{circle_node} = [circle, draw, minimum size=2em]
\tikzstyle{square_node} = [rectangle, draw, fill=red!20, minimum size=2em]

\node[circle_node] (Root) {$\circlearrowleft$};
\node[square_node] (A1) [below left of=Root] {$a_1$};
\node[square_node] (A2) [below right  of=Root] {$a_2$};

\draw[->] (Root) -- (A1);
\draw[->] (Root) -- (A2);

\end{tikzpicture}
}
      \end{tabular}& 
      \begin{tabular}{c}\\[-3mm]\resizebox{0.475\textwidth}{!}{
\begin{tikzpicture}[node distance=.5cm,>=stealth',thick]


\node[place,tokens=1] (p0) {};
\node[transition] (t0) [right=of p0, fill=black,label=below:$\tau$] {};

\node[place] (p1) [right =of  t0] {};

\node[transition] (t1) [right=of p1, label=center:$a_1$] {};
\node[place] (p1end) [right=of t1] {};

\node[place] (p2end) [below= 0.1cm of p1end] {};

\node[transition] (t2) [left=of p2end, label=center:$a_2$] {};



\node[transition] (tend) [right=of p1end, fill=black,label=below:$\tau$] {};
\node[transition] (tloop) [right =of p2end, fill=black,label=below:$\tau$] {};

\node[place] (pend) [right =of tend]{};

\draw[->] (p0) -- (t0);
\draw[->] (t0) -- (p1);

\draw[->] (p1) -- (t1);
\draw[->] (t1) -- (p1end);
\draw[->] (t2) -- (p1);
\draw[->] (p2end) -- (t2);

\draw[->] (p1end) -- (tend);
\draw[->] (p1end) -- (tloop);
\draw[->] (tloop) -- (p2end);
\draw[->] (tend) -- (pend);

\end{tikzpicture}
} 
      \end{tabular}\\
    \hline     
    \end{tabular}
    \caption{Mapping rules to obtain a WN from a PT}
    \label{fig:ptwn}
\end{figure}

\textit{Mapping of process tree into a workflow net}. 
PTs are naturally translated into corresponding WNs by application of the mapping rules depicted in Figure~\ref{fig:ptwn}. 
It is straightforward to show that the mapping is sound in the sense that a PT and its corresponding WN mapping are \emph{trace equivalent} (they produce the same set of traces). 
We stress that the parallel ($\land$) and the loop ($\circlearrowleft$) PT operators insert extra silent transitions in the WN mapping. This may result in overly complex WNs, that is, in WNs equipped with many silent transitions which, as we show later on, may unnecessarily augment the complexity of the optimal parameter search for the stochastic interpretation of the WNs (i.e., to solve the stochastic process discovery problem). 

\begin{myremark}
\label{rem:causality}
Observe that the activity labeled transitions $a_1$, $a_2$ to $a_n$ in the WN  mapping of the PT parallel operator $\land(a_1,\ldots, a_n)$ are not \emph{structurally causally connected} (Figure~\ref{fig:wnmapping}). This essentially means that firing of transition $a_i$ has no effect on the ability of $a_j$ ($j\neq i$) to fire\footnote{More technically, a transition $t_j$ is causally connected to  a transition $t_i$ \emph{iff} it is possible that the firing of $t_i$ increments the marking of some input place of $t_j$.}. Because of the recursive syntax of PT (Definition~\ref{def:ptsyntax}) the same remark applies to the transitions of the different WNs corresponding to mapping of different parallel sub-tress. Therefore if $N_i=(P_i,T_i,F_i,\Sigma,\lambda_i,M_{0i})$  ($1\leq i\leq n$) is the WN mapping of the $i$-th PT operand  $Q_i$ of the parallel tree  $\land(Q_1,Q_2,\ldots ,Q_n)$ then any transition $t_{j_i}\in T_i$ is not causally connected with any transition $t_{j'_k}\in T_k$ with $k\neq i$. 
\end{myremark}


\begin{example}
Figure~\ref{fig:wnmapping} shows the block structured  WN mapping corresponding to the process tree $Q_1$ of Figure~\ref{fig:pt1}.
\begin{figure}
    \centering
    \includegraphics[width=0.8\linewidth]{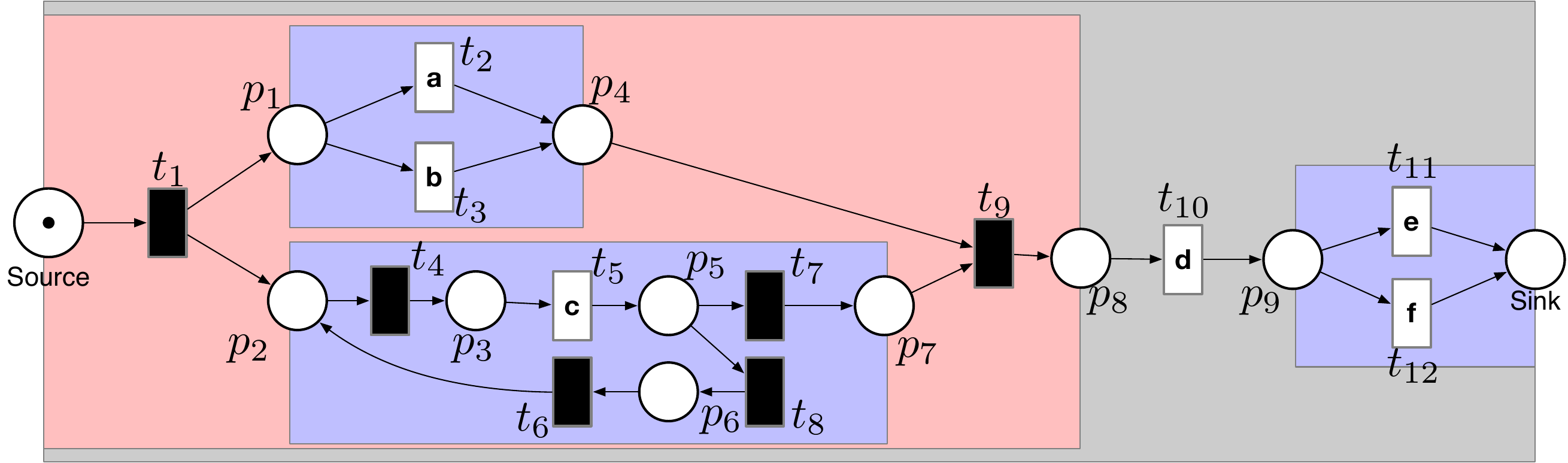}
    \caption{Block-structured WN resulting from the mapping  of the PT in Figure~\ref{fig:pt1}}
    \label{fig:wnmapping}
\end{figure}
Non-elementary WN  blocks are depicted within colored rectangles corresponding to the sub-trees of $Q_1$. Overall the WN consists  of 5 (black-filled) silent transitions ($t_1,t_4,t_6,t_7,t_8$)  and 6 activity transitions ($t_2,t_3,t_5,t_9,t_{10},t_{11}$). In the stochastic setting each transition $t_i$ is associated with a (positive real-valued) weight $w_i$ which determines the probability that $t_i$ occurs 
in any marking in which $t_i$ is concurrently enabled with other transitions. For example, in marking $p_1\!+\!p_2$ transitions $t_2,t_3,t_4$ are concurrently enabled therefore their respective weights determine with which probability one of them fires. For example, if $t_2$ fires then the next marking is $p_4\!+\!p_2$ and this happens with probability $Pr(p_1\!+\!p_2 [t_2\rangle p_4\!+\!p_2 )=w_2/(w_2+w_3+w_4)$. The product of such terms gives the probability that the token is moved from $Source$ to $Sink$ due to the firing of a given sequence of transitions. For example,
\begin{align} \nonumber
Pr(&Source
[t_1\rangle 
p_1\!+\!p_2 
[t_2\rangle 
p_4\!+\!p_2
[t_4\rangle 
p_4\!+\!p_3
[t_5\rangle \\& \nonumber
p_4\!+\!p_5
[t_7\rangle 
p_4\!+\!p_7
[t_9\rangle 
p_8
[t_{10}\rangle 
p_9
[t_{12}\rangle 
Sink)=\\& \label{eq:trseq}
\frac{w_2}{w_2\!+\!w_3\!+\!w_4}\cdot
\frac{w_7}{w_7\!+\!w_8}\cdot
\frac{w_{12}}{w_{11}\!+\!w_{12}}
\end{align}
The above sequence of transitions produces the trace
$\tr{a,c,d,f}$. Note however that (\ref{eq:trseq}) is not equal to $Pr(\tr{a,c,d,f})$ (i.e., the probability of producing the trace $\tr{a,c,d,f}$) because the trace $\tr{a,c,d,f}$ is generated by two sequences of transitions (because silent transition $t_4$ can fire before or after $t_2$). In general, several transition sequences can produce the same trace and, to calculate the probability of a trace, the probabilities of all transition sequences corresponding to the trace must be summed up. 
\end{example}
\section{A Motivating Example}
\label{sec:example}
Stochastic process discovery methods mostly follow an indirect approach, i.e., they use standard discovery algorithms (such as IM) to obtain a WN model from a log and then they work out adequate weight parameters for the transitions of the WN model, so that the resulting stochastic language resembles that of the log. We claim that in most cases such an approach results in WN models in which the number of parameters (weights) is unnecessarily large and their role in the corresponding stochastic language is not clear. This makes harder to find the optimal weights.

For this reason we introduce (in Section~\ref{sec:method}) the stochastic process tree formalism and propose to adapt indirect process discovery so that it operates directly on  stochastic process trees (SPTs) rather than on the corresponding WN model obtained through the mapping rules illustrated in Figure~\ref{fig:ptwn}. 

The advantage resulting from the SPT formalism is twofold: first we reduce the number of parameters that the corresponding SL depends upon and, second, the parameters have a clear role in the SL.
We illustrate this second aspect through a motivating example. 

Let us consider a popular, freely accessible event log, known as BPIC13\_open which contains observed executions of an incident management system of Volvo IT. The traces are built on an alphabet of three actions: \emph{Accepted} ($A$), \emph{Queued} ($Q$) and \emph{Completed} ($C$) which, respectively, refer to the acceptance, the queuing and the completion  of an incident query. 

\begin{figure}[t]
    \centering
    \resizebox{0.65\textwidth}{!}{
\begin{tikzpicture}[>=stealth, node distance=1.45cm, thick]

\tikzstyle{circle_node} = [circle, draw, minimum size=2em]
\tikzstyle{square_node} = [rectangle, draw, fill=red!20, minimum size=2em]

\node[circle_node] (Root) {$\land$};
\node[circle_node] (Choice11) [below left of=Root] {$\times$};

\node[circle_node] (Paral2) [below left of=Choice11] {$\land$};

\node[circle_node] (Choice2) [below right of=Root] {$\times$};

\node[circle_node] (Choice3) [below  of=Paral2] {$\times$};
\node[circle_node] (Choice4) [left of=Choice3] {$\times$};


\node[circle_node] (Loop2) [below right  of=Choice2] {$\circlearrowleft$};
\node[square_node] (Tau) [left  of=Loop2] {$\tau$};
\node[square_node] (A) [below  of=Loop2] {$A$};
\node[square_node] (Tau2) [right  of=A] {$\tau$};

\node[square_node] (Tau3) [right  of=Choice1] {$\tau$};

\node[square_node] (Tau5) [below   of=Choice4] {$\tau$};

\node[circle_node] (Loop3) [right  of=Tau5] {$\circlearrowleft$};

\node[circle_node] (Loop4) [left  of=Tau5] {$\circlearrowleft$};

\node[square_node] (Tau4) [right  of=Loop3] {$\tau$};

\node[square_node] (Tau6) [below   of=Loop3] {$\tau$};
\node[square_node] (C) [left   of=Tau6] {$C$};

\node[square_node] (Tau7) [left  of=C] {$\tau$};
\node[square_node] (Q) [left   of=Tau7] {$Q$};

\draw[->] (Root) -- (Choice11) node[midway, above left] {$p_1$};
\draw[->] (Root) -- (Choice2) node[midway, above right] {$\overline{p_1}$};
\draw[->] (Paral2) -- (Choice3) node[midway,  right] {$\overline{p_4}$};
\draw[->] (Paral2) -- (Choice4) node[midway,  left] {$p_4$};
\draw[->] (Choice11) -- (Paral2) node[midway,  left] {$p_2$};
\draw[->] (Choice11) -- (Tau3) node[midway,  right]  {$\overline{p_2}$};
\draw[->] (Loop2) -- (A) node[midway,  left] {$p_5$};
\draw[->] (Loop2) -- (Tau2)node[midway,  right] {$\overline{p_5}$};;

\draw[->] (Choice2) -- (Loop2) node[midway,  right] {$\overline{p_3}$};
\draw[->] (Choice3) -- (Tau4) node[midway,  right] {$\overline{p_7}$};
\draw[->] (Choice4) -- (Loop4)node[midway,  left] {$p_6$};
\draw[->] (Choice3) -- (Loop3) node[midway,  left] {$p_7$};
\draw[->] (Loop3) -- (Tau6) node[midway,  right] {$\overline{p_9}$};
\draw[->] (Loop3) -- (C) node[midway,  left] {$p_9$};

\draw[->] (Choice4) -- (Tau5) node[midway,  right] {$\overline{p_6}$};

\draw[->] (Choice2) -- (Tau) node[midway,  left] {$p_3$};

\draw[->] (Loop4) -- (Tau7) node[midway,  right] {$\overline{p_8}$};
\draw[->] (Loop4) -- (Q) node[midway,  left] {$p_8$};

\end{tikzpicture}
}
    \caption{The PT for   the BPIC13\_open  log obtained through the IM algorithm.}
    \label{fig:bpic13pt}
\end{figure}
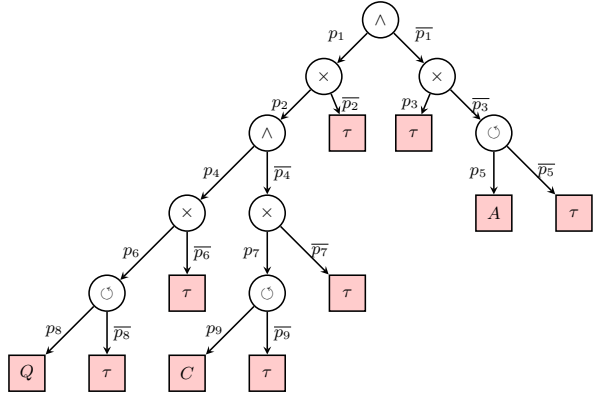

\begin{figure}[t]
    \centering    \includegraphics[width=0.85\linewidth]{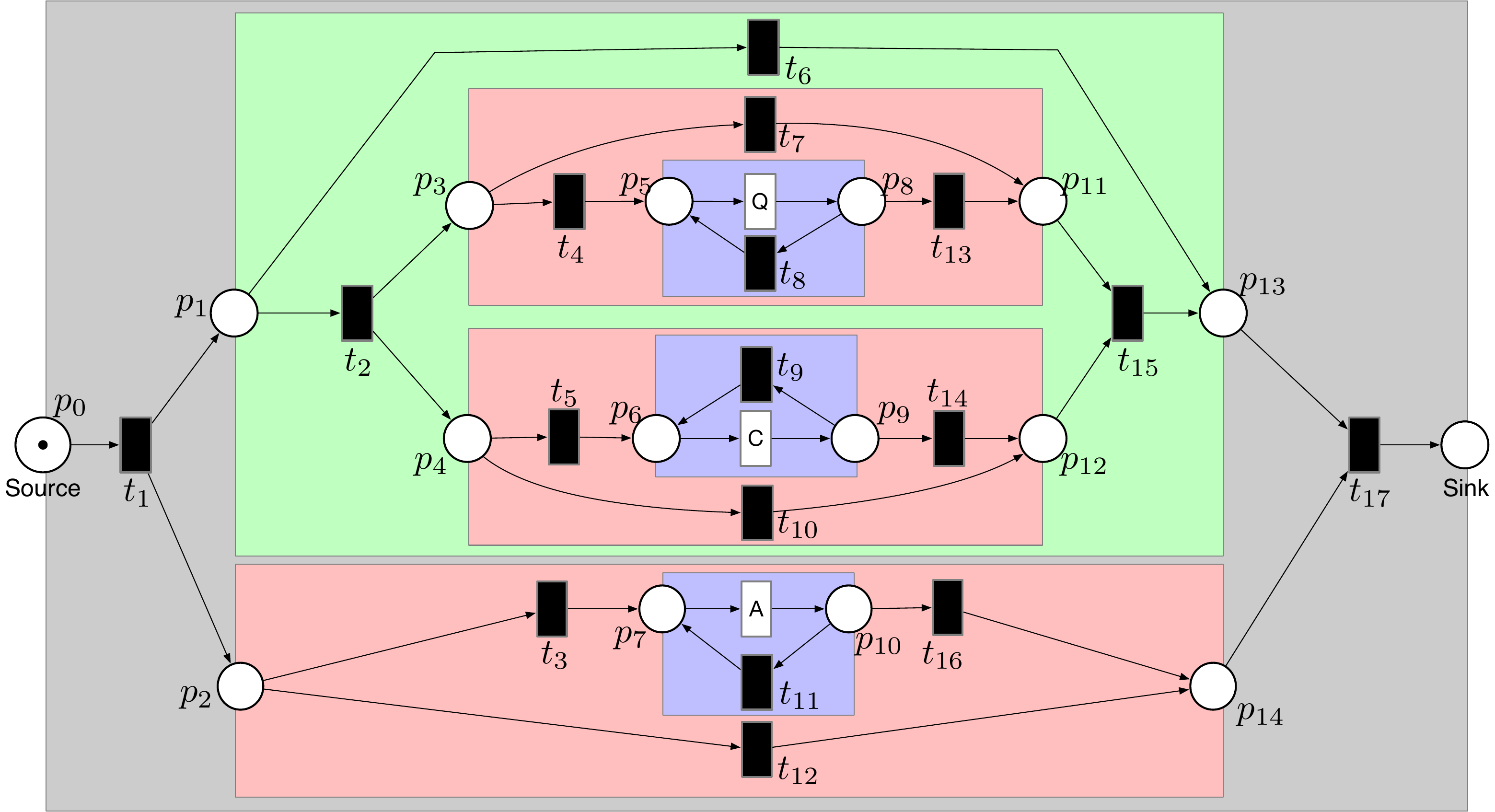}
    \caption{Block-structured WN obtained by conversion of the PT resulting from the BPIC13\_open log.}
    \label{fig:bpic13closedwn}
\end{figure}

Figure~\ref{fig:bpic13pt} shows the PT obtained by application of the IM discovery algorithm on the BPIC13\_open log, whereas Figure~\ref{fig:bpic13closedwn} depicts the corresponding WN translation. 
(The probabilities along the arcs in Figure~\ref{fig:bpic13pt} will be introduced and explained later.)

Looking at the block structure of  the WN in Figure~\ref{fig:bpic13closedwn} we observe that it consists of two main parallel blocks, the bottom (pink) one containing an optional $A$-loop block, the top (green) one further combining two parallel blocks containing an optional $Q$-loop and an optional $C$-loop block, respectively. This means the net is designed to issue words that may contain an undetermined number of $A$ actions (resulting from the bottom parallel block that contains the $A$-loop block) interleaved  with an undetermined number of shuffled  $Q$ and $C$ actions (resulting from the top block which contains the two parallel $Q$-loop and $C$-loop). 

Observe that the WN in Figure~\ref{fig:bpic13closedwn} consists of 20 transitions, 17 of which are silent while each of the remaining three corresponds to one of the three activities $A$, $C$ and $Q$. 
In the stochastic process discovery settings this means that an algorithm has to identify the optimal value for the 20 weight parameters and, in order to do so, it has to account for the impact that each transition weight has on the probability of each word of the stochastic language issued by the WN. We remind that the probability of a word (that is, a sequence of activities, in this case, e.g., $AQQQC$ or $ACAAQ$) is given by the sum of the probabilities of all those transition firing sequences that produce the considered word.
The problem in this respect is that, because of the likely presence of parallel blocks (to model concurrency)  and  of the possibly high number of silent transitions in the discovered WN, it turns out that many different firing sequences result in the same word (sequence of activities). From a  parameter optimisation perspective this leads to higher complexity, as the parameter of many (silent) transitions must be taken into account even in situations when, due to the absence of \emph{structural causality} (see Remark~\ref{rem:causality}), their value should not have a role in the probability of the resulting word. 

Let us illustrate this based on the WN discovered for the BPIC13\_open log (Figure~\ref{fig:bpic13closedwn}). Let us consider the word $w_1=AQC$. This word corresponds to many different firing sequences of the WN, for example, $t_1.t_2.t_3.A.t_5.\allowbreak t_4.Q.C.t_{13}.\allowbreak t_{14}.t_{15}.t_{16}.t_{17}$ and $t_1.t_2.t_3.A.t_4.\allowbreak t_5.Q.C.t_{13}.\allowbreak t_{14}.t_{15}.t_{16}.t_{17}$ and $t_1.t_2.t_3.t_4\allowbreak  .t_5.A.Q.C.\allowbreak t_{16}.\allowbreak t_{14}.t_{13}.t_{15}.t_{17}$, just to mention a few. Let us consider the first five firings in $t_1.t_2.t_3.A.t_5.\allowbreak t_4.Q.C.t_{13}.\allowbreak t_{14}.t_{15}.t_{16}.t_{17}$: 
\begin{equation*}
\begin{split}
    (p_0)[t_1\rangle (p_1\!+\! p_2) & [t_2\rangle (p_2\!+\! p_3\!+\! p_4) [t_3\rangle(p_3\!+\! p_4\!+\! p_7) [A\rangle (p_3\!+\! p_4\!+\! p_{10})[t_4\rangle\\
  & (p_4\!+\! p_5\!+\!p_{10})[t_5\rangle (p_5\!+\! p_6\!+\!p_{10})
\end{split}    
\end{equation*}
Notice that silent transition $t_{11}$ (responsible to re-activate transition $A$ hence to allow for multiple occurrences of $A$ in a trace) is enabled both in marking $(p_3\!+\! p_4\!+\! p_{10})$ and in $(p_5\!+\! p_6\!+\! p_{10})$. In $(p_3\!+\! p_4\!+\! p_{10})$ it is concurrently enabled with transitions $t_4$ and $t_5$ while in $(p_5\!+\! p_6\!+\! p_{10})$ with transitions $Q$ and $C$. There are several other markings in which $t_{11}$ is concurrently enabled with transitions of the blocks of $Q$ and $C$. This means that the weight of $t_{11}$ affects the probability of how activities $Q$ and $C$ are executed. Moreover, it affects it in several different ways according to the weights of the transitions it is concurrently enabled with. This, from a logical standpoint, is not ideal as transitions $Q$ and $C$ are part of blocks that are not causally connected with $t_{11}$. A large number of such situations can be found even in a relatively small example as those in Figure~\ref{fig:bpic13closedwn}. 

On the contrary let us consider the PT of the BPIC13\_open log in Figure~\ref{fig:bpic13pt} and let us imagine that each arc connecting a node with a children node is enriched with a probability parameter $p_i$ (parameters of the  arcs outgoing a given node must sum up to 1). We observe that there are nine probability parameters $p_i, 1\leq i\leq 9,$  which can be used to associate a probability value to each trace the SPT can generate (the formal semantics will be given in Section~\ref{sec:method}.)

Besides a considerable reduction in the number of parameters (9 probability parameters for the PT in Figure~\ref{fig:bpic13pt} as opposed to 20 weight parameters for the corresponding WN model in Figure~\ref{fig:bpic13closedwn}), which reduces the complexity of the optimization of the parameters, another advantage with the probability enriched PT model is that it avoids the inconvenience of having the probability of the words issued by a model affected by weight parameters of transitions which are causally non-connected (as discussed above). In other words, with the stochastic process tree formalism we introduce in Section~\ref{sec:method}, the role of a probability parameter associated with an operator will be {\em local} to the sub-tree whose root is its operator. As a consequence the parameters have a clear role in how they impact the SL issued by a process tree.

Another drawback about the idea of introducing stochasticity at WN level, rather than at  PT level (as done by all stochastic process discovery approaches based on IM), is due to the lack of univocality of the discovered WN. The IM algorithm, in fact, extracts a PT from a log and then produces a corresponding WN through mapping rules (see Figure~\ref{fig:ptwn}). We stress, however, that the resulting WN is not necessarily the only possible representation of the PT extracted by the IM algorithm. It is straightforward to show that several, semantically equivalent WN models can be obtained, for example, as modifications of that resulting from the mapping rules.

\begin{figure}
    \centering    \includegraphics[width=0.85\linewidth]{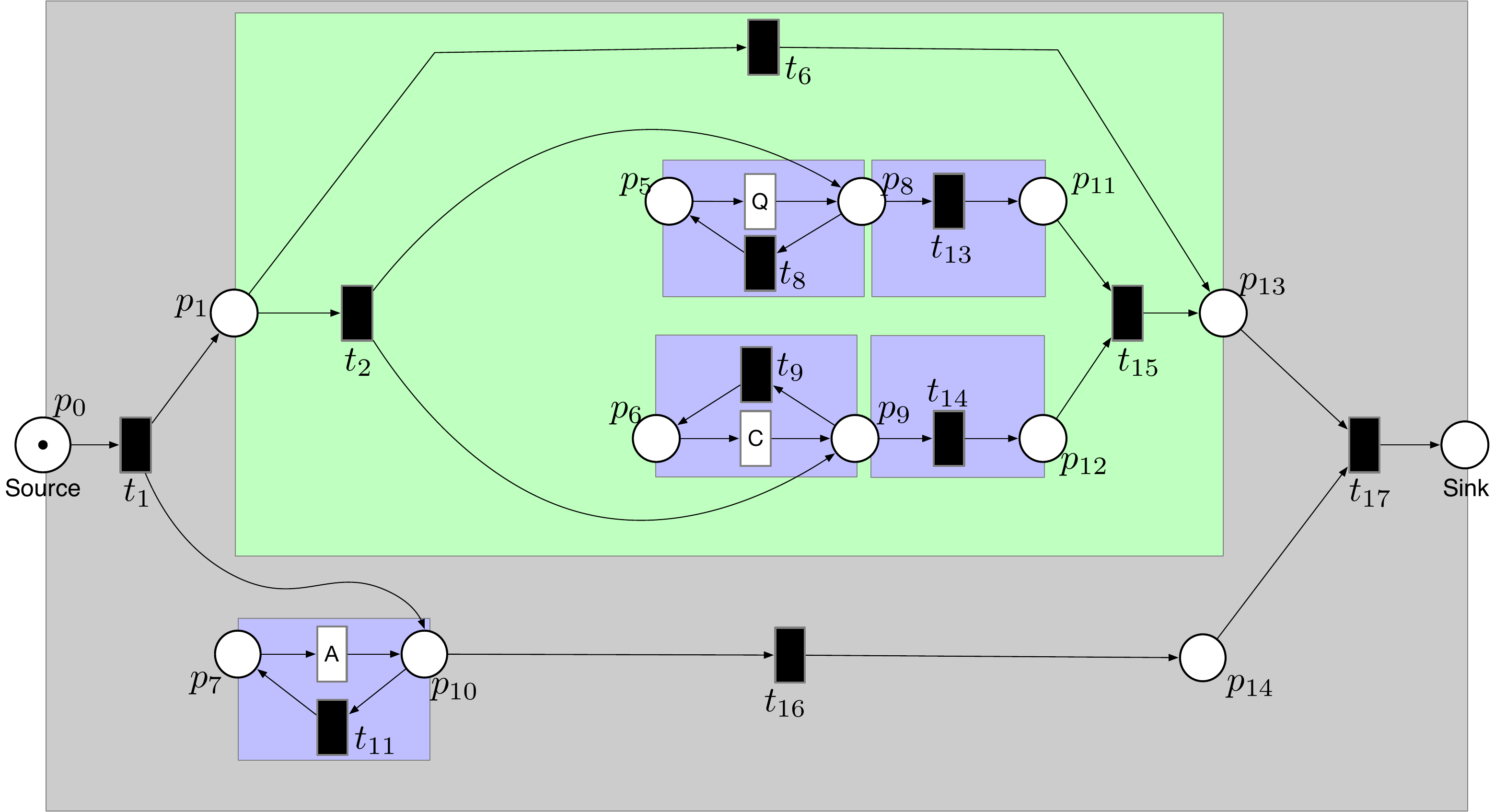}
    \caption{A semantically equivalent WN to that of Figure~\ref{fig:bpic13closedwn}.}
    \label{fig:bpic13closedwn2}
\end{figure}

For example, Figure~\ref{fig:bpic13closedwn2} shows a WN  obtained by removing a few transitions and places from that
resulting from the mapping rules applied by the IM algorithm and depicted in Figure~\ref{fig:bpic13closedwn}. It is straightforward to show that these two WNs are semantically equivalent (i.e., they produce the same language) however the net in Figure~\ref{fig:bpic13closedwn2} is clearly less complex and has significantly fewer transitions (14 as opposed to 20). 

As a consequence we stress that we can have several different WNs with different number of parameters that obviously behave differently when it comes to parameter optimization. 
 
Therefore by introducing the stochastic extension of the process tree formalism we avoid this inconvenience by allowing for the optimisation step to operate directly at the process tree level rather than on (not univocal) WN models. 

In the next section we formalize the notion of stochastic process tree (SPT). Together with the syntax we formalize how the stochastic language issued by an SPT is determined in function of the probability parameters of the involved operators. Furthermore, we provide a procedure   for sampling words belonging to the  stochastic language corresponding to a given SPT, as well as, a procedure to obtain an approximation of the stochastic language of an SPT. 

For the sake of completeness, we mention here \cite{ChMaBaCo93} where the authors deal with issues related to the role of the weights associated with immediate transitions in Petri nets with both timed and immediate transitions.  The paper suggests that applying true concurrency semantics to ``non-conflicting’' transitions could lead in some cases to a reduction of the number of parameters.  Although potentially relevant, their findings cannot be straightforwardly applied in our context.  In fact, when it comes to determining the language issued by a WN, we need to account for the fact that transitions are either labeled or silent and that, even in the case of non-conflicting transitions, the firing order has an impact on the language (except  in the case of concurrently enabled non-conflicting silent transitions). Moreover, mined WNs likely contain many markings in which transitions belonging to different effective conflict sets are concurrently enabled.


\section{Stochastic Process Trees}
\label{sec:method}

A process tree is associated with a language, that is, a set of traces it represents. In order to deal with stochastic languages, in which every trace is associated with a probability, we extend process trees with probabilistic mechanisms. 
In the following we  first provide the definition of stochastic process trees (SPT), then we intuitively illustrate the meaning of the involved stochastic operators  through simple examples before formally defining  their semantics.

\begin{definition}[Stochastic process tree]\label{def:sptsyntax} 
Let $\Sigma$ be an alphabet of process activities and $\{\rightarrow, \times, \land, \circlearrowleft  \}$ the set of stochastic process tree operators. The set of SPTs over $\Sigma$ is recursively defined as:
\begin{itemize}
\item if $a \in\Sigma \cup \{\tau\}$ then $Q=a$ is a SPT,        
\item if $Q_1,...,Q_n$ with $n \geq 2$ are SPTs then $\rightarrow(Q_1,Q_2,...,Q_n)$ is a SPT,
\item if $Q_1,...,Q_n$ with $n \geq 2$ are SPTs, $0 \leq p_i \leq 1, i=1,2,...n$ with $\sum_{i=1}^n p_i=1$, and $\oplus \in \{\times, \land \}$ then $\oplus(Q_1,...,Q_n,p_1,...,p_n)$ is a SPT,
\item if $Q_1$ and $Q_2$ are SPTs and $0\leq p \leq 1$ then $\circlearrowleft(Q_1,Q_2,p)$ is a SPT.
\end{itemize}
\end{definition}

In order to illustrate the semantics of the four operators we consider  three SPTs $Q_1$, $Q_2$ and $Q_3$, with corresponding stochastic languages:
\begin{equation}
\label{eq:slex}
    L_1(\tr{a,b})=\frac{1}{4},~L_1(\tr{b,a})=\frac{3}{4},~L_2(\tr{c})=1,~L_3(\tr{d})=\frac{1}{2},~L_3(\tr{e})=\frac{1}{2}.
    \end{equation}

The simplest case is the sequence operator. Consider, as an example,
$Q=\rightarrow(Q_1,Q_2,Q_3)$ which implies that a trace of $Q$ is generated by concatenation of three traces generated by $Q_1$, $Q_2$ and $Q_3$, respectively. Denoting the SL of $Q$ by $L=\mathcal{L}(Q)$, we have 
\begin{align*}
L(\tr{a,b,c,d})=L_1(\tr{a,b}) \cdot L_2(\tr{c}) \cdot
L_3(\tr{d})=\frac{1}{4}\!\cdot\!\frac{1}{2}=\frac{1}{8}
\end{align*}
while the other possibilities are with probabilities
\begin{align*}&
L(\tr{a,b,c,e})=\frac{1}{4}\!\cdot\!\frac{1}{2}=\frac{1}{8},~
L(\tr{b,a,c,d})=L(\tr{b,a,c,e})=\frac{3}{4}\!\cdot\!\frac{1}{2}=\frac{3}{8}
\end{align*}
where the total probability, as expected, is $2\!\cdot\!\frac{1}{8}+2\!\cdot\!\frac{3}{8}=1$.

The choice operator, $\times(Q_1,...,Q_n,p_1,...,p_n)$, is made stochastic by assigning a probability $p_i$ to tree $Q_i$, where $p_i$ represents the probability of selecting a trace from $Q_i$. In other words, the SL of $\times(Q_1,...,Q_n,p_1,...,p_n)$ is the mixture of the SLs of $Q_1,Q_2,...,Q_n$ with weights $p_1,p_2,...,p_n$.
As an example, let us consider $Q_1$ and $Q_3$ as defined before with probabilities $p_1=\frac{1}{5}$ and $p_3=\frac{4}{5}$. The SL  $L=\mathcal{L}(\times(Q_1,Q_3,p_1,p_3))$ is composed of the traces
\begin{align*} &
L(\tr{a,b})=p_1\cdot L_1(\tr{a,b})=\frac{1}{5}\!\cdot\!\frac{1}{4}=\frac{1}{20},~
L(\tr{b,a})=p_1\cdot L_1(\tr{b,a})=\frac{1}{5}\!\cdot\!\frac{3}{4}=\frac{3}{20},\\&
L(\tr{d})=p_2\cdot L_2(\tr{d})=\frac{4}{5}\!\cdot\!\frac{1}{2}=\frac{2}{5},~
L(\tr{e})=p_2\cdot L_2(\tr{e})=\frac{4}{5}\!\cdot\!\frac{1}{2}=\frac{2}{5}
\end{align*}
where $\frac{1}{20}+\frac{3}{20}+2\!\cdot\!\frac{2}{5}=1$. 

The parallel operator, $\land(Q_1,...,Q_n,p_1,...,p_n)$, is complemented with $n$ probability values $p_i$ that, simply speaking, drive from which of the components languages $\mathcal{L}(Q_i)$ the next activity is chosen in forming, by concatenation, the traces of the resulting language. 

The key point about parallel composition is that traces of the resulting language $\mathcal{L}(
\land(Q_1,...,Q_n,p_1,...,p_n))$ are obtained by \emph{shuffling} of (the actions of the) traces of the component languages. Intuitively, the probability of a parallel composed word $\sigma$ resulting by shuffling of $n$ words $\sigma_i\in\mathcal{L}(Q_i)$ ($1\leq i\leq n$) can be thought of as being determined as follows: first each component trace $\sigma_i$ participating in the shuflle is selected from the corresponding language $\mathcal{L}(Q_i)$ with its corresponding probability $\mathcal{L}(Q_i)(\sigma_i)$, then the shuffled word $\sigma$ is determined  by  concatenating the actions of the participant traces $\sigma_i$ using the probabilities $p_i, 1\leq i \leq n$ to select from which trace the next action is taken and added into the forming shuffle trace. 

For example, let us consider the parallel composition of languages of  $Q_1$ and $Q_2$ using shuffling  probabilities $p_1=\frac{1}{3}$ and $p_2=\frac{2}{3}$. The resulting SL $L=\mathcal{L}(\land(Q_1,Q_2,p_1,p_2))$ is given by
\begin{align*}&
L(\tr{c,a,b})=\frac{1}{4}\!\cdot\!\frac{2}{3}=\frac{1}{6},~
L(\tr{c,b,a})=\frac{3}{4}\!\cdot\!\frac{2}{3}=\frac{1}{2}, \\&
L(\tr{a,c,b})=\frac{1}{4}\!\cdot\!\frac{1}{3}\!\cdot\!\frac{2}{3}=\frac{1}{18},~
L(\tr{b,c,a})=\frac{3}{4}\!\cdot\!\frac{1}{3}\!\cdot\!\frac{2}{3}=\frac{1}{6}, \\&
L(\tr{a,b,c})=\frac{1}{4}\!\cdot\!\frac{1}{3}\!\cdot\!\frac{1}{3}=\frac{1}{36},~
L(\tr{b,a,c})=\frac{3}{4}\!\cdot\!\frac{1}{3}\!\cdot\!\frac{1}{3}=\frac{1}{12},
\end{align*}
which sums up to 1. 
Let us discuss, for example, how the probability of the shuffled word $\sigma=\langle a,c,b\rangle$ is determined. Trace $\sigma=\langle a,c,b\rangle$ is the result of shuffling of trace $\sigma_1=\langle a,b\rangle\in\mathcal{L}(Q_1)$, which is selected with probability $\mathcal{L}(Q_1)(\langle a,b\rangle)=\frac{1}{4}$, and trace $\sigma_2=\langle c\rangle\in\mathcal{L}(Q_2)$, which is selected with probability $\mathcal{L}(Q_2)(\langle c\rangle)=1$.  The resulting shuffle $\langle a,c,b\rangle$ indicates that the first action $a$ is taken from trace  $\langle a,b\rangle\in\mathcal{L}(Q_1)$, which happens with probability $p_1=\frac{1}{3}$, while the second action $c$ is taken from $\langle c\rangle\in\mathcal{L}(Q_2)$, which happens with probability $p_2=\frac{2}{3}$. Finally the last action $b$, being  the only remaining one, is chosen, with probability 1, from $\sigma_1$. Therefore the resulting probability of shuffle $\langle a,c,b\rangle$ is given by the product of each of the above mentioned probability values, that is: $L(\tr{a,c,b})=\frac{1}{4}\!\cdot\!1\!\cdot\! \frac{1}{3}\!\cdot\!\frac{2}{3}\!\cdot\!1=\frac{1}{18}$.

As shown in the example discussed above, we point out a peculiarity about  the characterisation of the probability of shuffled words, that is: the probability values $p_i$ have to be accounted for (in selecting the next action to be added in the shuffle) only up until letters of the corresponding word $\sigma_i$ remain to be added in the shuffle. As soon as all letters of $\sigma_i$ have been ``consumed'', $p_i$ stops to contribute to the probabilistic selection of the next action to be added 
and, therefore, the probabilistic selection continues  via normalisation of the probability values $p_j$ of the remaining (unconsumed) traces $\sigma_j$. 
This is captured by the formal characterisation of the probability of  traces of the resulting language  $\mathcal{L}(
\land(Q_1,...,Q_n,p_1,...,p_n))$  defined  in~(\ref{eq:shuffling}). 

The loop operator, $\circlearrowleft(Q_1,Q_2,p)$, comes with a single parameter $p$ which represents the probability of looping back. As in case of the non-stochastic case (Definition \ref{def:ptsemanitcs}), $Q_1$ is used at least once. Then at each turn, looping back happens with probability $p$, concatenating first a trace from $Q_2$ and then another from $Q_1$. If looping back does not happen (which has probability $1-p$) the trace is ready.


Let us consider a few concrete elements of the SL issued by the SPT given as $Q=\circlearrowleft(Q_2,Q_1,p)$ with $Q_2$ and $Q_1$ as defined in~(\ref{eq:slex})  and looping probability $p=\frac{2}{5}$. The corresponding SL, $\mathcal{L}(Q)$, is with
\begin{align*}
\mbox{1 execution:}~~~&L(\tr{c})=L_1(\tr{c})\cdot (1-p)=\frac{3}{5}, \\
\mbox{2 executions:}~~~&L(\tr{c,a,b,c})=
L_1(\tr{c})\cdot p \cdot L_2(\tr{a,b})\cdot L_1(\tr{c})\cdot (1-p)=
\frac{3}{50},\\
&L(\tr{c,b,a,c})=
L_1(\tr{c})\cdot p \cdot L_2(\tr{b,a})\cdot L_1(\tr{c})\cdot (1-p)=
\frac{9}{50}, \\
\mbox{3 executions:}~~~&
L(\tr{c,a,b,c,a,b,c})=\frac{2}{5}\!\cdot\!\frac{1}{4}
\!\cdot\!\frac{2}{5}\!\cdot\!\frac{1}{4}
\!\cdot\!\left(1-\frac{2}{5}\right)=\frac{3}{500},\\ &
L(\tr{c,a,b,c,b,a,c})=\frac{2}{5}\!\cdot\!\frac{1}{4}
\!\cdot\!\frac{2}{5}\!\cdot\!\frac{3}{4}
\!\cdot\!\left(1-\frac{2}{5}\right)=\frac{9}{500},\\ &
L(\tr{c,b,a,c,a,b,c})=\frac{2}{5}\!\cdot\!\frac{3}{4}
\!\cdot\!\frac{2}{5}\!\cdot\!\frac{1}{4}
\!\cdot\!\left(1-\frac{2}{5}\right)=\frac{9}{500},\\ &
L(\tr{c,b,a,c,b,a,c})=\frac{2}{5}\!\cdot\!\frac{3}{4}
\!\cdot\!\frac{2}{5}\!\cdot\!\frac{3}{4}
\!\cdot\!\left(1-\frac{2}{5}\right)=\frac{27}{500},\\ 
\mbox{4 executions:}~~~ & ...
\end{align*}
where for 3 executions we listed only numerical values to save some space.
It is evident that, unless $p=0$, the SL of a loop operator comes with an infinite number of traces. In the above list, traces with $i$ executions have a total probability of $p^{(i-1)}\!\cdot\!(1-p)$ and, since $\sum_{i=1}^\infty p^{i-1}\!\cdot\!(1-p)=1$, we obtain a normalized SL.

A SPT induces a SL over the words of the language of its embedded PT. 
\begin{definition}[Semantics of SPTs]\label{def:sptsemantics} 
Let $Q$ be a SPT. The SL $L=\mathcal{L}(Q)$ of $Q$ is recursively defined as follows (where $L_i=\mathcal{L}(Q_i)$ denotes the stochastic language of sub-tree $Q_i$):
\begin{itemize}
\item if $Q=a$ with $a \in\Sigma$ then $L(\tr{a})=1$,  
\item if $Q=\tau$ then $L(\tr{})=1$,  
\item (sequence:) if $Q=\rightarrow(Q_1,Q_2,...,Q_n)$ then
\begin{align*}
L(\sigma)=\sum_{\forall \sigma_1.\sigma_2.~...~.\sigma_n=\sigma}
\prod_{i=1}^n L_{i}(\sigma_{i}),
\end{align*}
\item (choice:) if $Q=\times(Q_1,...,Q_n,p_1,...,p_n)$ then
\begin{align*}
L(\sigma)=\sum_{i=1}^n p_i L_i(\sigma),
\end{align*}
\item (parallel:) if $Q=\land(Q_1,...,Q_n,p_1,...,p_n)$ then
\begin{align*}
L(\sigma)=\sum_{\forall \sigma_1,\sigma_2,...,\sigma_n}
P(Sh(\sigma_1,\sigma_2,...,\sigma_n,p_1,p_2,...,p_n)=\sigma)
\prod_{i=1}^n L_i(\sigma_i),
\end{align*}
where $P(Sh(\sigma_1,...,\sigma_n,p_1,...,p_n)=\sigma)$ is the probability that the stochastic shuffling of $\sigma_1,\sigma_2,...,\sigma_n$ using the probabilities $p_1,p_2,...,p_n$ results in $\sigma$, whose definition is given by Equation~(\ref{eq:shuffling}). 
\item (loop:) if $Q=\circlearrowleft(Q_1,Q_2,p)$ then
\begin{align*}
L(\sigma)=\sum_{\substack{m=1,2,3,... \\ \forall \sigma_1.\sigma_1'.\sigma_2.\sigma_2'.~...~.\sigma_{m-1}'.\sigma_m=\sigma}}
p^{(m-1)}(1-p) \prod_{i=1}^m L_1(\sigma_i) \prod_{i=1}^{m-1} L_2(\sigma_i').
\end{align*}
\end{itemize}
\end{definition}

In order to calculate the probability that stochastic shuffling of traces $\sigma_1,...,$ $\sigma_n$ using probabilities $p_1,...,p_n$ results in $\sigma$, we introduce some notation. A shuffling can be described by a sequence $s=[i_1,i_2,...,i_{|\sigma|}]$ with  $1\leq i_j \leq n, 1\leq j \leq |\sigma|$ representing the fact that the $k$-th element of $\sigma$ comes from $\sigma_{s[k]}$. The result of shuffling traces $\sigma_1, \sigma_2, \ldots ,\sigma_n$ according to $s$ will be denoted by $R(\sigma_1,\sigma_2,...,\sigma_n,s)$. 
For example, if $\sigma_1=\tr{a,a,b}$, $\sigma_2=\tr{c,d}$ and $s=[1,1,2,2,1]$ then $R(\sigma_1,\sigma_2,s)=\tr{a,a,c,d,b}$. 
Given a shuffling $s$, we denote by $\mathcal{C}(s,k)$ the set of indices of those traces that have not yet been included completely in $\sigma$ before position $k$.
With our previous example, $\mathcal{C}(s,1)=\mathcal{C}(s,2)=\mathcal{C}(s,3)=\mathcal{C}(s,4)=\{1,2\}$ and $\mathcal{C}(s,5)=\{1\}$. In other words, $\mathcal{C}(s,k)$ indicates the set of traces that can still contribute in position $k$. Further, let $\mathcal{S}(\sigma_1,...,\sigma_n,\sigma)$ denote the set of all possible shufflings of traces $\sigma_1,...,\sigma_n$ that result in $\sigma$, that is, $\mathcal{S}(\sigma_1,...,\sigma_n,\sigma)=\{s|R(\sigma_1,...,\sigma_n,s)=\sigma\}$. With this notation, the probability that the stochastic shuffling of $\sigma_1,\sigma_2,...,\sigma_n$ using probabilities $p_1,p_2,...,p_n$ results in $\sigma$ can be written as
\begin{align}
\label{eq:shuffling}
P(Sh(\sigma_1,\sigma_2,...,\sigma_n,p_1,p_2,...,p_n)=\sigma)=
\sum_{s\in\mathcal{S}(\sigma_1,...,\sigma_n,\sigma)}
\prod_{k=1}^{|\sigma|} \frac{p_{s[k]}}{\sum_{i \in \mathcal{C}(s,k)} p_i}.
\end{align}
where the product gives the probability that the shuffling is according to $s$.
In our previous example, if we use shuffling probabilities $p_1=\frac{1}{3}$ and $p_2=\frac{2}{3}$ then we have, for example,
\begin{align*}&
P(Sh(\sigma_1,\sigma_2,p_1,p_2)=\tr{a,a,c,d,b})=\frac{1}{3}\!\cdot\!\frac{1}{3}\!\cdot\!\frac{2}{3}\!\cdot\!\frac{2}{3}\!\cdot\!1=\frac{4}{81}, \\ &
P(Sh(\sigma_1,\sigma_2,p_1,p_2)=\tr{c,d,a,a,b})=\frac{2}{3}\!\cdot\!\frac{2}{3}\!\cdot\!1\!\cdot\!1\!\cdot\!1=\frac{4}{9}.
\end{align*}

\begin{myremark}
In Definition \ref{def:sptsemantics} the operators $\rightarrow\mbox{ (sequence)}, \land\mbox{ (parallel)}$ and $\circlearrowleft\mbox{ (loop)}$ are defined through summations over a set of potentially very large number of traces. In many practical situations, for example, when the involved SPTs are with disjoint alphabets, the summations can become much simpler. We leave the identification and exploitation of these situations to future work.
\end{myremark}

\begin{myremark}
The sequence operator, $\rightarrow(Q_1,Q_2,...,Q_n)$, can also be defined in such a way that the order in which traces are taken from the component trees is stochastic. We avoided this in order to not overload the definition with parameters ($n$ trees can be arranged in a sequence in $n!$ different ways). Random order of the component trees can be realized by combining the choice and the sequence operators. For example, in case of $\times(\rightarrow(Q_1,Q_2,Q_3),\rightarrow(Q_3,Q_2,Q_1),\frac{1}{3},\frac{2}{3})$, the order in which the trees are used is $Q_1,Q_2,Q_3$ with probability $\frac{1}{3}$ and it is $Q_3,Q_2,Q_1$ with probability $\frac{2}{3}$.   
\end{myremark}

\begin{myremark}
The loop operator is such that the number of executions follows a geometric distribution. This choice is driven by the fact that this way the underlying stochastic process is memoryless and lends itself better to efficient non-simulation-based analysis, a topic we wish to develop in future papers. If the event log suggests that the number of executions of an activity or a sequence of activities is not properly modeled by a geometric distribution, the formalism can be easily extended to other distributions.
\end{myremark}

Generation of a trace by simulation given a SPT can be carried out in a recursive fashion starting from the root of the tree. 
\begin{procedure}[Simulation of SPTs]
\label{proc:sptsim}
The recursive procedure to obtain a trace $\sigma$ from $Q$ by simulation is the following:
\begin{itemize}
\item if $Q=a$ with $a \in\Sigma$ then $\sigma=\tr{a}$,  
\item if $Q=\tau$ then $\sigma=\tr{}$,  
\item if $Q=\rightarrow(Q_1,Q_2,...,Q_n,O)$ then 
\begin{enumerate}
    \item generate traces $\sigma_1,...,\sigma_n$ according to $Q_1,...,Q_n$
    \item by concatenation $\sigma=\sigma_{1}.\sigma_{2}.~...~.\sigma_{n}$ 
\end{enumerate}
\item if $Q=\times(Q_1,...,Q_n,p_1,...,p_n)$ then
\begin{enumerate}
    \item choose $i$ from $\{1,...,n\}$ according to the probabilities $p_1,...,p_n$
    \item $\sigma$ is a trace generated from $Q_i$
\end{enumerate}
\item if $Q=\land(Q_1,...,Q_n,p_1,...,p_n)$ then
\begin{enumerate}
    \item generate traces $\sigma_1,...,\sigma_n$ according to $Q_1,...,Q_n$
    \item set $i_1=1,i_2=1,...,i_n=1$ and $\sigma=\tr{}$
    \item choose $j$ from $\{1,...,n\}$ according to the probabilities $p_1,...,p_n$
    \item if $|\sigma_j| \geq i_j$ set $\sigma=\sigma.\sigma_j[i_j]$ and $i_j=i_j+1$
    \item if $\land_{j=1}^n i_j>|\sigma_j|$ return $\sigma$ otherwise go back to step 3
\end{enumerate}
\item if $Q=\circlearrowleft(Q_1,Q_2,p)$ then
\begin{enumerate}
    \item choose the number of executions $m$ according to the geometric distribution $P(m=i)=p^{i-1}(1-p)$ with $i\geq 1$
    \item generate $m$ traces $\sigma_1,...,\sigma_m$ according to $Q_1$
    \item generate $m-1$ traces $\sigma_1',...,\sigma_{m-1}'$ according to $Q_2$
    \item by concatenation $\sigma=\sigma_1.\sigma_1'.\sigma_2.\sigma_2'.~...~.\sigma_{m-1}'.\sigma_{m}$ 
\end{enumerate}
\end{itemize}
\end{procedure}

\begin{myremark}
From a computational point of view step 3 of case $Q=\land(Q_1,...,Q_n$, $p_1,...,p_n)$ can be improved, in order to not to waste random numbers, by choosing $j$ only among those traces that have not yet been included in $\sigma$ completely. Here we preferred this simpler description.
\end{myremark}

If a tree contains at least one loop operator, $\circlearrowleft(Q_1,Q_2,p)$, with parameter $p>0$ then the corresponding SL is composed of an infinite number of traces.  A finite approximation of such a language can be computed by the following procedure where the maximal number of cycles is limited by $C_{\max}$ for all loop operators of the tree. In an implementation of the procedure the language can be  represented by a set of pairs containing a trace and the associated probability.  We use the notation $L(\sigma)+\!\!=p$ meaning that if there is already a pair with trace $\sigma$ then its probability is increased by $p$, otherwise a pair with $\sigma$ and probability $p$ is added to the set of pairs.

\begin{procedure}[Approximate SL of a SPT]\label{proc:approxL}
Given a stochastic process tree, $Q$, an approximation of the  associated stochastic language can be calculated as 
\begin{itemize}
\item if $Q=a$ with $a \in\Sigma$ then $L(\tr{a})=1$,  
\item if $Q=\tau$ then $L(\tr{})=1$,  
\item if $Q=\rightarrow(Q_1,Q_2,...,Q_n,O)$ then
\begin{align*}
& \forall \sigma_1~\mbox{s.t.}~L_1(\sigma_1)>0, \forall \sigma_2~\mbox{s.t.}~L_2(\sigma_2)>0,..., \forall \sigma_n~\mbox{s.t.}~L_n(\sigma_n)>0,\\
&~~~L(\sigma_{1}.\sigma_{2}.~...~.\sigma_{n})+\!\!= \prod_{i=1}^n L_i(\sigma_i)
\end{align*}
\item if $Q=\times(Q_1,...,Q_n,p_1,...,p_n)$ then
\begin{align*}
& \forall 1\leq i \leq n, \forall \sigma~\mbox{s.t.}~L_i(\sigma)>0,\\ 
&~~~ L(\sigma)+\!\!=p_i\cdot L_i(\sigma)
\end{align*}
\item if $Q=\land(Q_1,...,Q_n,p_1,...,p_n)$ then
\begin{align*}
& \forall \sigma_1~\mbox{s.t.}~L_1(\sigma_1)>0, \forall \sigma_2~\mbox{s.t.}~L_2(\sigma_2)>0,..., \forall \sigma_n~\mbox{s.t.}~L_n(\sigma_n)>0,\\
& \forall~\mbox{shuffling}~s~\mbox{of traces of length}~|\sigma_1|,|\sigma_2|,...,|\sigma_n|, \\
& ~~~L(R(\sigma_1,\sigma_2,...,\sigma_n,s))+\!\!=
\left(\prod_{k=1}^{|\sigma_1|+...+|\sigma_n|} \frac{p_{s[k]}}{\sum_{i \in \mathcal{C}(s,k)} p_i}\right)\left( \prod_{i=1}^n L_i(\sigma_i)\right)
\end{align*}
where the first product is the probability that shuffling $s$ is chosen. All possible shufflings can be obtained instead by considering all vectors $s$ of length $|\sigma_1|+...+|\sigma_n|$ in which the number of entries equal to $j$ is $|\sigma_j|$.
\item if $Q=\circlearrowleft(Q_1,Q_2,p)$ then
\begin{align*}
& \forall 1\leq m \leq C_{\max}, \\
& \forall \sigma_1,\sigma_2,...,\sigma_m~\mbox{s.t}~L_1(\sigma_1)>0,L_1(\sigma_2)>0,...,L_1(\sigma_m)>0,\\
& \forall \sigma_1',\sigma_2',...,\sigma_{m-1}'~\mbox{s.t}~L_2(\sigma_1')>0,L_2(\sigma_2')>0,...,L_2(\sigma_{m-1}')>0,\\
&~~~L(\sigma_1.\sigma_1'.\sigma_2.\sigma_2'.~...~.\sigma_{m-1}'.\sigma_m)+\!\!=
p^{(m-1)}(1-p) \prod_{i=1}^m L_1(\sigma_i) \prod_{i=1}^{m-1} L_2(\sigma_i').
\end{align*}
\end{itemize}
\end{procedure}
\begin{myremark}
It is straightforward to modify Procedure \ref{proc:approxL} in such a way that each loop operator of the tree is controlled by its own maximal number of loops, or by a probabilistic constraint that guarantees that the considered number of loops cover a given amount of probability.
\end{myremark}

\section{Preliminary Experiments}
\label{sec:exper}

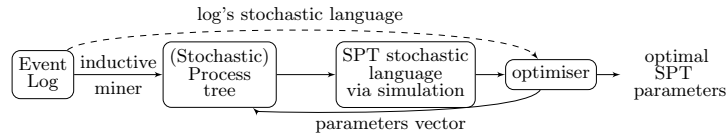
\begin{figure}
    \centering
\resizebox{.8\textwidth}{!}{
\begin{tikzpicture}[scale=.6,minimum width=1cm]
\renewcommand{\arraystretch}{.7}
   \everymath{\scriptstyle}
   
\node[left, initial text=,draw,rounded corners] (l0) { \begin{tabular}{@{}c@{}}\text{Event}\\  \text{Log} \end{tabular}};
\node[right=1.5cm of l0,draw,rounded corners] (l1) { \begin{tabular}{@{}c@{}}\text{(Stochastic)}\\ \text{Process}\\  \text{tree} \\  \end{tabular} };

\node[right=1cm of l1,draw,rounded corners] (l3) { \begin{tabular}{@{}c@{}}\text{SPT stochastic} \\ \text{language}  \\  \text{via   simulation}
\end{tabular} };

\node[right=.5cm of l3,draw,rounded corners] (l4) { \begin{tabular}{@{}c@{}}\text{optimiser}\\ 
\end{tabular} };
\node[right=.4cm of l4] (l5){\begin{tabular}{@{}c@{}}\text{optimal}\\  \text{SPT } \\ 
\text{ parameters} \\
\end{tabular}};

\draw [-latex'] (l0) -- (l1) node [midway, above,sloped] {inductive} node [midway, below,sloped] {miner};
\draw [-latex'] (l1) -- (l3) node [midway, above,sloped] { };
\draw [-latex'] (l3) -- (l4) node [midway, above,sloped] { };
\draw [-latex'] (l4) .. controls +(235:15mm) and +(315:15mm) .. (l1) node [pos=0.5 ,below] {parameters vector};
\draw [-latex'] (l4) -- (l5) node [midway, above,sloped] {};
\draw [-latex',dashed] (l0) .. controls +(45:20mm) and +(130:20mm) .. (l4) node [pos=0.5 ,above] {log's stochastic language};

{}; 
 
\end{tikzpicture}
}
    \caption{Stochastic process discovery based on SPT simulation.}
    \label{fig:spttool}
\end{figure}

In order to test the SPT formalism introduced in Section~\ref{sec:method} we have implemented a prototype software tool whose functioning is outlined in Figure~\ref{fig:spttool}. 

The tool takes an event log $E$ as input and extracts (by application of  IM discovery) the corresponding PT  which is then enriched with adequate probability parameters $p_i$ to obtain the corresponding SPT $Q(E)$. 
The parameter values are initially  randomly chosen and then progressively modified through an iterative optimizer that searches for the optimal values minimizing the restricted rEMD distance. 
Specifically, in our implementation, the rEMD 
is computed by considering the event log $E$ itself as the finite subset of the (possibly infinite) language $\mathcal{L}(Q(E))$ issued by  $Q(E)$\footnote{Notice that, since IM has exact fitness, it is guaranteed that $E\subseteq \mathcal{L}(Q(E))$.}. 


The tool depends on a number of configuration parameters which includes the number of iterations after which the optimizer stops and the number of SPT traces sampled by the simulator (used to approximate the SPT stochastic language). Intuitively, both of these configuration parameters affect the outcome produced by the tool, in the sense that the larger the number of optimization iterations the better  the optimization and the larger the number of sampled traces the better the approximation of the SPT language (hence, indirectly the quality  of the optimization). 


\begin{table}
    \centering
\resizebox{.9\textwidth}{!}{    \begin{tabular}{l p{1ex} c c  c c  c l p{1ex} c c c c} 
        \toprule 
        Event log & &  \multicolumn{2}{l}{log properties} & & \multicolumn{1}{l}{WN} & & \multicolumn{1}{l}{SPT} & & \multicolumn{4}{c}{SPT-based optimisation} \\
        \cmidrule(){3-4} \cmidrule(){6-6} \cmidrule(){8-8} \cmidrule(){10-13}
         & & \#traces & \#act & & \#pars &  & \#pars & & Starting vector rEMD & \#iter  & rEMD & runtime (s) \\ \hline
        \multirow{5}{*}{BPIC13\_c} & & \multirow{5}{*}{183} & \multirow{5}{*}{4} & & \multirow{5}{*}{20} & & \multirow{5}{*}{10} & & \multirow{5}{*}{0.21} & 1 & 0.17 &  \\
        & & & & & & & & &  & 2 & 0.12 &  \\
        & & & & & & & & &   & 5 & 0.09 &  \\
        & & & & & & & & &   & 10 & 0.08 &  \\
        & & & & & & & & &   & 20 & 0.07 & 223 \\
        \hline
        \multirow{5}{*}{BPIC13\_i} & & \multirow{5}{*}{1511} & \multirow{5}{*}{4} & & \multirow{5}{*}{23} & &  \multirow{5}{*}{11} & & \multirow{5}{*}{0.3} & 1 & 0.15 &  \\
        & & & & & & & & &   & 2 & 0.13 &  \\
        & & & & & & & & &   & 5 & 0.1 &  \\
        & & & & & & & & &   & 10 & 0.08 &  \\
        & & & & & & & & &   & 20 & 0.05 & 196 \\
        \hline
        \multirow{5}{*}{BPIC13\_o} & &  \multirow{5}{*}{108} &  \multirow{5}{*}{3} & & \multirow{5}{*}{19} & &  \multirow{5}{*}{10} & & \multirow{5}{*}{0.32} & 1 & 0.227 &  \\
        & & & & & & & & &  & 2 & 0.225 &  \\
        & & & & & & & & &  & 5 & 0.221 & \\
        & & & & & & & & &   & 10 & 0.21 &  \\
        & & & & & & & & &   & 20 & 0.2 & 195 \\
        \bottomrule
    \end{tabular}
    \caption{Optimised SPT-based process discovery on  BPIC13 event logs.}
    \label{tab:exp1}
    }
\end{table}


Table~\ref{tab:exp1} reports about experiments obtained by running the SPT-based process discovery software tool on three versions of a real-systems event log taken from the \emph{Business Process Intelligence Challenge} (BPIC). Specifically, we considered three logs of the 2013 BPIC~\cite{DBLP:ceur-ws.org/Vol-1052}, denoted BPIC13\_c, BPIC13\_i and BPIC13\_o  (as in  \emph{closed}, \emph{incidents}, respectively \emph{open})  provided by the information technology department of Volvo car manufacturer in Belgium. 
The logs contain events from an incident and an (open and closed) problem management system aimed at restoring a customer’s normal service operation as quickly as possible when
incidents or a problem arise. 

The \emph{closed}, \emph{incident} and \emph{open} logs contain, respectively, 183, 1511 and 108 traces composed  over alphabets of respectively 4, 4 and 3 actions. Column  ``$\sharp$pars'' in Table~\ref{tab:exp1}  depicts the number of parameters for the  discovered WN and SPT model, which highlights that SPT essentially halves the dimension of the parameter space with respect to using a Petri net model obtained from a PT. Column ``Starting vector rEMD'' indicates the rEMD distance between the language of the log and the language obtained using a randomly chosen, initial vector of SPT parameters. The progression of the optimization process is provided in column ``rEMD'' in function of the number of iterations (column ``$\sharp$iter``) of the optimization engine, which indicates asymptotic convergence towards optimal distance. Finally, execution time, in seconds, of each experiment is reported in column ``runtime'' for a total of 20 iterations.

We run similar experiments using the PT based approach with results reported in \cite{DBLP:journals/corr/abs-2406-10817}. The final rEMD values for the three considered logs in the same order as here were 0.04, 0.18 and 0.07. Meaning that in two cases out of three the PT based experiments provided better results. Note however, that for this paper we used a preliminary simulation based implementation which gives less guaranties that the optimal values are found. In any case the final rEMD values in Table~\ref{tab:exp1} show that the models provide similar distributions to that of the log. 

As future work we plan to implement an efficient computational algorithm to calculate the language of an SPT (similar to the unfolding used in \cite{DBLP:journals/corr/abs-2406-10817} for the PT based approach). We expect that this will improve the final rEMD results and decrease significantly the execution time.  
\section{Conclusion}
\label{sec:conclusion}

When stochastic models are sought for describing processes of organizations, the standard approach proceeds by, first, finding a non-stochastic representation (most often a process tree), second, transforming it to a Petri net and, last, adding stochasticity to the Petri net. In this paper we illustrated that this procedure has unfavorable properties, such as, having to deal with a non well-defined number of parameters with ambiguous roles in the resulting stochastic process. 

As one source of these problems is the transformation of process trees into Petri nets, in this paper, we proposed a formalism, namely, stochastic process trees, that allows for adding stochasticity directly to process trees. Besides the syntax and semantics of stochastic process trees, we provided two procedures to work with them: the first one for the simulation while the second for the approximate calculation of the stochastic language of a stochastic process tree.

We also performed preliminary numerical experiments based on simulation using event logs of real systems. These suggest that stochastic process trees have a similar precision in capturing the stochastic aspects of event logs as Petri net models obtained from process trees. At the same time, stochastic process trees have the advantage of having a fewer number of parameters, which is advantageous when it comes to optimization, and the role of the parameters is easier to interpret.

As future work we plan to develop efficient analysis algorithms for stochastic process trees and carry out a broad set of experiments on real event logs in order to demonstrate the validity of the approach.


%
%
%
%

\bibliographystyle{plain} 
\bibliography{biblio}

\begin{thebibliography}{10}

\bibitem{10.1007/s10115-018-1214-x}
Adriano Augusto, Raffaele Conforti, Marlon Dumas, Marcello La~Rosa, and Artem
  Polyvyanyy.
\newblock Split miner: automated discovery of accurate and simple business
  process models from event logs.
\newblock {\em Knowl. Inf. Syst.}, 59(2):251–284, may 2019.

\bibitem{DBLP:conf/icpm/BurkeLW20}
Adam Burke, Sander J.~J. Leemans, and Moe~Thandar Wynn.
\newblock Stochastic process discovery by weight estimation.
\newblock In Sander J.~J. Leemans and Henrik Leopold, editors, {\em Process
  Mining Workshops - {ICPM} 2020 International Workshops, Padua, Italy, October
  5-8, 2020, Revised Selected Papers}, volume 406 of {\em Lecture Notes in
  Business Information Processing}, pages 260--272. Springer, 2020.

\bibitem{DBLP:conf/apn/BurkeLW21}
Adam Burke, Sander J.~J. Leemans, and Moe~Thandar Wynn.
\newblock Discovering stochastic process models by reduction and abstraction.
\newblock In Didier Buchs and Josep Carmona, editors, {\em Application and
  Theory of Petri Nets and Concurrency - 42nd International Conference, {PETRI}
  {NETS} 2021, Virtual Event, June 23-25, 2021, Proceedings}, volume 12734 of
  {\em Lecture Notes in Computer Science}, pages 312--336. Springer, 2021.

\bibitem{ChMaBaCo93}
Giovanni Chiola, Marco Marsan, Gianfranco Balbo, and Gianni Conte.
\newblock Generalized stochastic petri nets: A definition at the net level and
  its implications.
\newblock {\em IEEE Trans. Software Eng.}, 19:89--107, 02 1993.

\bibitem{DBLP:journals/corr/abs-2406-10817}
Pierre Cry, Andr{\'a}s Horv{\'a}th, Paolo Ballarini, and Pascale Le~Gall.
\newblock A framework for optimisation based stochastic process discovery.
\newblock In Jane Hillston, Sadegh Soudjani, and Masaki Waga, editors, {\em
  Quantitative Evaluation of Systems (QEST) and Formal Modeling and Analysis of
  Timed Systems (FORMATS)}, volume 14996 of {\em LNCS}, pages 34--51, Cham,
  2024. Springer Nature Switzerland.

\bibitem{DBLP:conf/apn/LeemansFA13}
Sander J.~J. Leemans, Dirk Fahland, and Wil M.~P. van~der Aalst.
\newblock Discovering block-structured process models from event logs - {A}
  constructive approach.
\newblock In Jos{\'{e}}~Manuel Colom and J{\"{o}}rg Desel, editors, {\em
  Application and Theory of Petri Nets and Concurrency - 34th International
  Conference, {PETRI} {NETS} 2013, Milan, Italy, June 24-28, 2013.
  Proceedings}, volume 7927 of {\em Lecture Notes in Computer Science}, pages
  311--329. Springer, 2013.

\bibitem{10.1007/978-3-030-26643-1_8}
Sander J.~J. Leemans, Anja~F. Syring, and Wil M.~P. van~der Aalst.
\newblock Earth movers' stochastic conformance checking.
\newblock In Thomas Hildebrandt, Boudewijn~F. van Dongen, Maximilian
  R{\"o}glinger, and Jan Mendling, editors, {\em Business Process Management
  Forum}, pages 127--143, Cham, 2019. Springer International Publishing.

\bibitem{pearson1894}
Karl Pearson.
\newblock Contributions to the mathematical theory of evolution.
\newblock {\em Philosophical Transactions of the Royal Society of London.
  Series A}, 185:71--110, 1894.

\bibitem{5f0e8dd04572478fb450eefe4211d69f}
A.~Rogge-Solti, W.M.P. {Aalst, van der}, and M.H. Weske.
\newblock Discovering stochastic petri nets with arbitrary delay distributions
  from event logs.
\newblock In N.~Lohmann, M.~Song, and P.~Wohed, editors, {\em Business Process
  Management Workshops : BPM 2013 International Workshops, Beijing, China,
  August 26, 2013, Revised Papers}, Lecture Notes in Business Information
  Processing, pages 15--27, Germany, 2014. Springer.
\newblock 9th International Workshop on Business Process Intelligence (BPI
  2013), BPI 2013 ; Conference date: 26-08-2013 Through 26-08-2013.

\bibitem{rubner1998signature}
Y.~Rubner, C.~Tomasi, and L.J. Guibas.
\newblock A metric for distributions with applications to image databases.
\newblock In {\em Proceedings of the IEEE Conference on Computer Vision and
  Pattern Recognition (CVPR)}, pages 59--66, 1998.

\bibitem{1316839}
W.~van~der Aalst, T.~Weijters, and L.~Maruster.
\newblock Workflow mining: discovering process models from event logs.
\newblock {\em IEEE Transactions on Knowledge and Data Engineering},
  16(9):1128--1142, 2004.

\bibitem{10.5555/2948762}
Wil van~der Aalst.
\newblock {\em Process Mining: Data Science in Action}.
\newblock Springer Publishing Company, Incorporated, 2nd edition, 2016.

\bibitem{https://doi.org/10.1002/widm.1045}
Wil van~der Aalst, Arya Adriansyah, and Boudewijn van Dongen.
\newblock Replaying history on process models for conformance checking and
  performance analysis.
\newblock {\em WIREs Data Mining and Knowledge Discovery}, 2(2):182--192, 2012.

\bibitem{10.1007/978-3-642-34044-4_3}
Wil van~der Aalst, Joos Buijs, and Boudewijn van Dongen.
\newblock Towards improving the representational bias of process mining.
\newblock In Karl Aberer, Ernesto Damiani, and Tharam Dillon, editors, {\em
  Data-Driven Process Discovery and Analysis}, pages 39--54, Berlin,
  Heidelberg, 2012. Springer Berlin Heidelberg.

\bibitem{DBLP:ceur-ws.org/Vol-1052}
Wil M.~P. van~der Aalst, Sander J.~J. Leemans, and Arthur H.~M. ter Hofstede,
  editors.
\newblock {\em Proceedings of the 1st International Workshop on Algorithms \&
  Theories for the Analysis of Event Data, {ATAED} 2013, Beijing, China, August
  26-30, 2013}, volume 1052 of {\em {CEUR} Workshop Proceedings}. CEUR-WS.org,
  2013.

\bibitem{10.1007/11494744_25}
B.~F. van Dongen, A.~K.~A. de~Medeiros, H.~M.~W. Verbeek, A.~J. M.~M. Weijters,
  and W.~M.~P. van~der Aalst.
\newblock The prom framework: A new era in process mining tool support.
\newblock In Gianfranco Ciardo and Philippe Darondeau, editors, {\em
  Applications and Theory of Petri Nets 2005}, pages 444--454, Berlin,
  Heidelberg, 2005. Springer Berlin Heidelberg.

\end{thebibliography}



\end{document}